\documentclass{article}

\usepackage[final,nonatbib]{ctrlgen_neurips_2021}

\usepackage{array}
\usepackage{colortbl}
\usepackage{graphicx}
\usepackage[utf8]{inputenc} 
\usepackage[T1]{fontenc}    
\usepackage{hyperref}       
\usepackage{url}            
\usepackage{booktabs}       
\usepackage{amsfonts}       
\usepackage{nicefrac}       
\usepackage{microtype}      
\usepackage{xcolor}         
\usepackage{multirow}     
\usepackage{amsmath}
\usepackage[colorinlistoftodos,prependcaption,textsize=tiny]{todonotes}
\usepackage{enumitem}

\newcommand{\mcc}[3]{\multicolumn{#1}{#2}{#3}}
\definecolor{Gray}{gray}{0.90}
\newcolumntype{a}{>{\columncolor{Gray}}r}
\newcolumntype{j}{>{\columncolor{Gray}}l}
\newcolumntype{g}{>{\columncolor{Gray}}c}
\newcommand{\good}[1]{\textcolor{blue}{#1}}
\newcommand{\bad}[1]{\textcolor{red}{#1}}

\newcommand{\framework}{{\sc Luminous}}
\newcommand{\alfred}{ALFRED}
\newcommand{\thor}{AI2Thor}

\newcommand{\bedroom}{{\it Bedroom}}
\newcommand{\bathroom}{{\it Bathroom}}
\newcommand{\livingroom}{{\it Living Room}}
\newcommand{\kitchen}{{\it Kitchen}}

\newcommand{\splits}[1]{{\it #1}}

\title{
LUMINOUS: Indoor Scene Generation\\for Embodied AI Challenges}

\author{%
  Yizhou Zhao$^1$\thanks{University of California, Los Angeles. Correspondence to:
Yizhou Zhao <yizhouzhao@g.ucla.edu> or Kaixiang Lin <kaixianglin.cs@gmail.com>}
  \And
  Kaixiang Lin$^2$
  \And
  Zhiwei Jia$^{3}$
  \And
  Qiaozi Gao$^2$
  \AND
  Govind Thattai$^2$
  \And
  Jesse Thomason$^{2,4}$
  \And
  Gaurav S. Sukhatme$^{2,4}$ 
  \AND
  \textnormal{$^1$University of California, Los Angeles; $^2$Amazon Alexa AI; }
  \AND
  \textnormal{$^3$University of California, San Diego;  $^4$University of Southern California}
}

\begin{document}

\maketitle

\begin{abstract}
Learning-based methods for training embodied agents typically require
a large number of high-quality scenes that contain realistic layouts
and support meaningful interactions.  However, current simulators for
Embodied AI (EAI) challenges only provide simulated indoor scenes with
a limited number of layouts.  This paper presents \framework, the
first research framework that employs state-of-the-art indoor scene
synthesis algorithms to generate large-scale simulated scenes for
Embodied AI challenges.  Further, we automatically and quantitatively
evaluate the quality of generated indoor scenes via their ability to
support complex household tasks.  \framework\ incorporates a novel
scene generation algorithm (Constrained Stochastic Scene Generation
(CSSG)), which achieves competitive performance with human-designed
scenes.  Within \framework, the EAI task executor, task instruction
generation module, and video rendering toolkit can collectively
generate a massive multimodal dataset of new scenes for the training
and evaluation of Embodied AI agents.  Extensive experimental results
demonstrate the effectiveness of the data generated by \framework,
enabling the comprehensive assessment of embodied agents on
generalization and robustness. 
The full codebase and documentation of
\framework\ is available at:
\url{https://github.com/amazon-research/indoor-scene-generation-eai/}.
\end{abstract}

\section{Introduction}

Embodied artificial intelligence (EAI) has attracted significant attention, both in advanced deep learning models and algorithms~\cite{vaswani2017attention,lu2019vilbert,suglia2021embodied,zhang2021hierarchical} and 
the rapid development of simulated platforms~\cite{puig2018virtualhome,kolve2017ai2,gan2020threedworld,li2021igibson,savva2019habitat}. 
Many open challenges~\cite{shridhar2020alfred,szot2021habitat,RoomR,shen2020igibson} have been proposed to facilitate EAI research. 
A critical bottleneck in existing simulated platforms~\cite{shridhar2020alfred,RoomR,li2021igibson,puig2018virtualhome,yan2018chalet} is the limited number of indoor scenes that support vision-and-language navigation, object interaction, and complex household tasks. This limitation makes
it difficult to verify whether state-of-the-art methods generalize well to unseen scenarios or whether they are specialized to a small number of room structures.
Low cost, automatic creation of large numbers of high-quality simulated environments is essential to resolve this question.

Here, we leverage advances in
indoor scene synthesis to achieve the large-scale automatic creation of simulated environments. 
Indoor scene synthesis has been a long-standing challenge
for both computer graphics and machine learning communities resulting in considerable recent progress~\cite{yu2011make,fisher2012example,fisher2015activity,qi2018human,wang2019planit,luo2020end,zhou2019scenegraphnet,wang2020sceneformer,zhang2020deep}.
To effectively utilize
indoor scene synthesis for EAI, three key challenges remain. First, for synthesized scenes to be useful in EAI, they must  directly support household tasks requiring object pick and place, state changes, and articulation. Second, the generated scenes with randomized layouts must be \textit{natural}---layouts that ``make sense'' according to human judgement---and \textit{functional}---layouts that match human use given the room type, such as \bedroom\ or \livingroom.
Finally, any scene generation method must provide efficient access to massive, multimodal embodied agent trajectory data, including  low-level action sequences for task completion, egocentric image frames during action execution, and language instructions.

\begin{figure}[t!]
    \centering
    \includegraphics[width=0.98\textwidth]{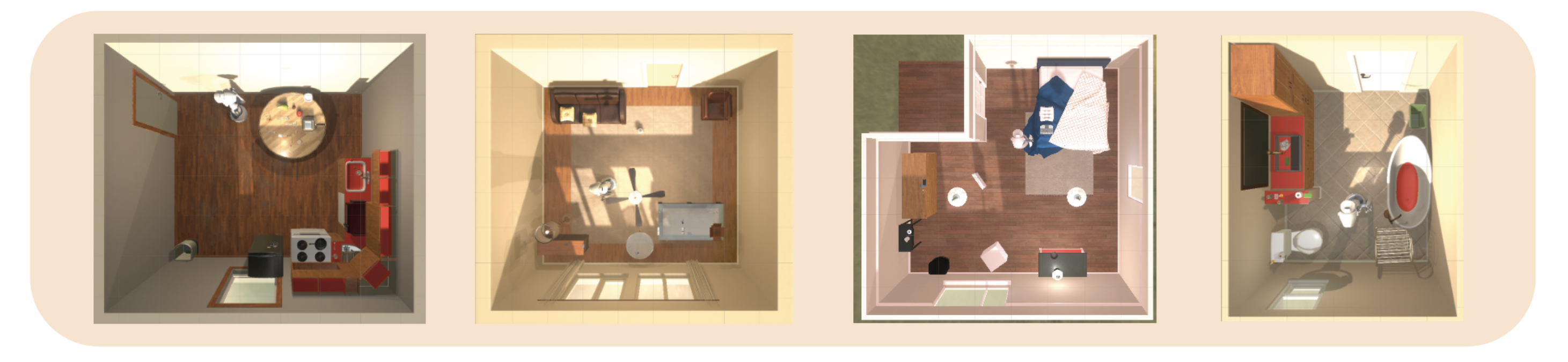}
    \caption{\textbf{Generated Indoor scenes.} \framework\ scenes are evaluated quantitatively via EAI task success rates and qualitatively via human judgements.}
    \label{fig:title_image}
\end{figure}

We present \framework, a scalable, indoor scene generation framework to facilitate EAI
tasks such as vision-and-language navigation and language-guided task
completion (Figure~\ref{fig:title_image}). We introduce the Challenge Definition Format (CDF), 
which provides a user-friendly task specification of the required objects, their relative spatial
relationships, and high-level descriptions of downstream EAI 
tasks to facilitate. We introduce Constrained Stochastic Scene Generation (CSSG)
to generate an arbitrary number of indoor scenes from the CDF specification. \framework\ produces scenes that are well-aligned with human common sense and
satisfy the CDF conditions, thereby ensuring that the generated scenes are readily applicable to EAI tasks. In addition, we develop a task solver to plan sequences of low-level actions for
corresponding task completion. We also implement a task instruction generation module to annotate trajectories with language instructions. 
\framework\ generates large-scale multimodal trajectories for the training and evaluation of embodied agents. 

\framework\ also contributes to indoor scene
synthesis. 
Generally, scene generation lacks ground truth for
quantitative evaluation. 
Metrics like bounding box and angle prediction~\cite{luo2020end} and synthetic
classification~\cite{wang2019planit} are not always correlated
with the quality of a generated scene. By connecting indoor scene synthesis to EAI, we propose measuring planner-based task success rate as an automatic evaluation metric of the synthesized scene quality. 
Besides CSSG, \framework\ is compatible with state-of-the-art learning-based indoor scene synthesis algorithms~\cite{wang2018deep,luo2020end}.
We demonstrate that CSSG with \framework\ qualitatively outperforms other learning-based synthesis methods (Section~\ref{sec:exp:scenequality}).

The main contributions of our work are threefold. 
First, we introduce a framework (\framework) which serves as a standard and unified benchmark for indoor scene synthesis algorithms. 
Second, \framework\ generates a large number of randomized scenes that achieve competitive quality compared to human-designed scenes in \thor~\cite{kolve2017ai2}. Third, the rendered scenes, along with the multimodal
trajectories, directly support typical EAI task completion to facilitate generalization research. 
Extensive evaluation on \alfred~\cite{shridhar2020alfred}, a language-guided task completion challenge,
demonstrate the effectiveness and scalability of \framework. 
Further, our evaluation with \framework\ scenes suggests that existing, state of the art models for \alfred\ may overfit to the hand-created scenes in \thor.

\section{Related Work}

\framework\ builds on and extends research in indoor scene synthesis, simulation environments in EAI, and language-guided task completion.

\textbf{Indoor Scene Synthesis}.
In computer graphics, extensive research exists in 3D indoor scene synthesis.  
Early work either used explicit rule-based constraints~\cite{xu2002constraint}
or incorporated stochastic priors into the generative
procedure~\cite{yu2011make,fisher2012example,fisher2015activity,qi2018human}.
Recent advances~\cite{wang2019planit,luo2020end,wang2020sceneformer} utilize
deep neural networks to extract patterns from large-scale
datasets~\cite{song2017semantic}. 
While these data-driven approaches significantly enhance the automation of the scene generation process, the resulting synthesized scenes
are still relatively simple in terms of object quantity and inter-object spatial
relationships.  
Many works generate scenes based on the natural representation of the scene graph~\cite{zhou2019scenegraphnet,wang2019planit,luo2020end}. 
Other lines of research condition on the image~\cite{wang2018deep,ritchie2019fast} or
text~\cite{ma2018language,chang2014semantic} representation of indoor scenes.
The discrepancies in the input representation of scene generation models and the
diverse sources of data make it difficult to compare and contrast the performance of different methods.  To facilitate research in 
learning-based approaches, \framework\ is designed to support end-to-end scene generation evaluation and a unified rendering tool to accommodate the outputs of various approaches simultaneously. 

\textbf{Embodied AI Simulators}. 
In the past few years, researchers have developed many simulation environments~\cite{kolve2017ai2,gan2020threedworld,shen2020igibson,puig2018virtualhome,savva2019habitat}
 to serve as training and evaluation platforms for embodied agents. These simulation environments propel research progress in a wide
range of embodied tasks, including vision-and-language task
completion~\cite{shridhar2020alfred,singh2020moca},
rearrangement~\cite{RoomR,gan2020threedworld},
navigation~\cite{savva2019habitat,shen2020igibson},
manipulation~\cite{xiang2020sapien,james2020rlbench} and human-robot
collaboration~\cite{puig2018virtualhome}.  Recently,
AllenAct~\cite{weihs2020allenact} integrates a set of embodied environments
(such as iThor, RoboThor, Habitat~\cite{savva2019habitat}, etc.), tasks, and
algorithms thereby facilitating the evaluation of the
same model or algorithm across multiple EAI platforms. Many EAI platforms
are designed with sophisticated indoor scenes to perform embodied tasks. Platforms such as
iGibson~\cite{shen2020igibson}, AI2Thor~\cite{kolve2017ai2} can randomize materials,
color, and small objects in the scene, while the basic room layouts remain unchanged.
To facilitate more robust and thorough evaluation of embodied agents, 
\framework\ automatically generates indoor scenes with randomized layouts at a large scale that readily support vision-and-language navigation and high-level object interactions. We summarized the properties of \framework\ and most popular EAI simulation platforms in Table~\ref{tab:simulations}.

\textbf{Language-Guided Task Completion}.
Among existing EAI challenges, we use \alfred~\cite{shridhar2020alfred} as 
our downstream exemplar task to evaluate the scene generation quality of \framework. \alfred\ enables agents to follow natural language descriptions to complete complex household tasks. 
\alfred\ tasks involve resolving vision-and-language grounding, affordance-aware navigation, and high-level object interactions. Roughly speaking, there are two categories of approaches to tackling \alfred. Initial approaches learned end-to-end models that mapped language instructions into low-level actions directly~\cite{singh2020moca,suglia2021embodied,pashevich2021episodic}. 
Subsequently, hierarchical approaches~\cite{zhang2021hierarchical,blukis2021persistent} were proposed that enabled better generalization and interpretation. 
However, those approaches are only tested in four indoor scenes unseen during training time.
Towards a more convincing evaluation, 
\framework\ generates an order of magnitude larger number of scenes for better assessment of generalization and robustness. 

\section{\sc \textbf{Luminous}}
\begin{figure}[t!]
    \centering
    \hspace*{-1cm}\includegraphics[width=1.1\textwidth]{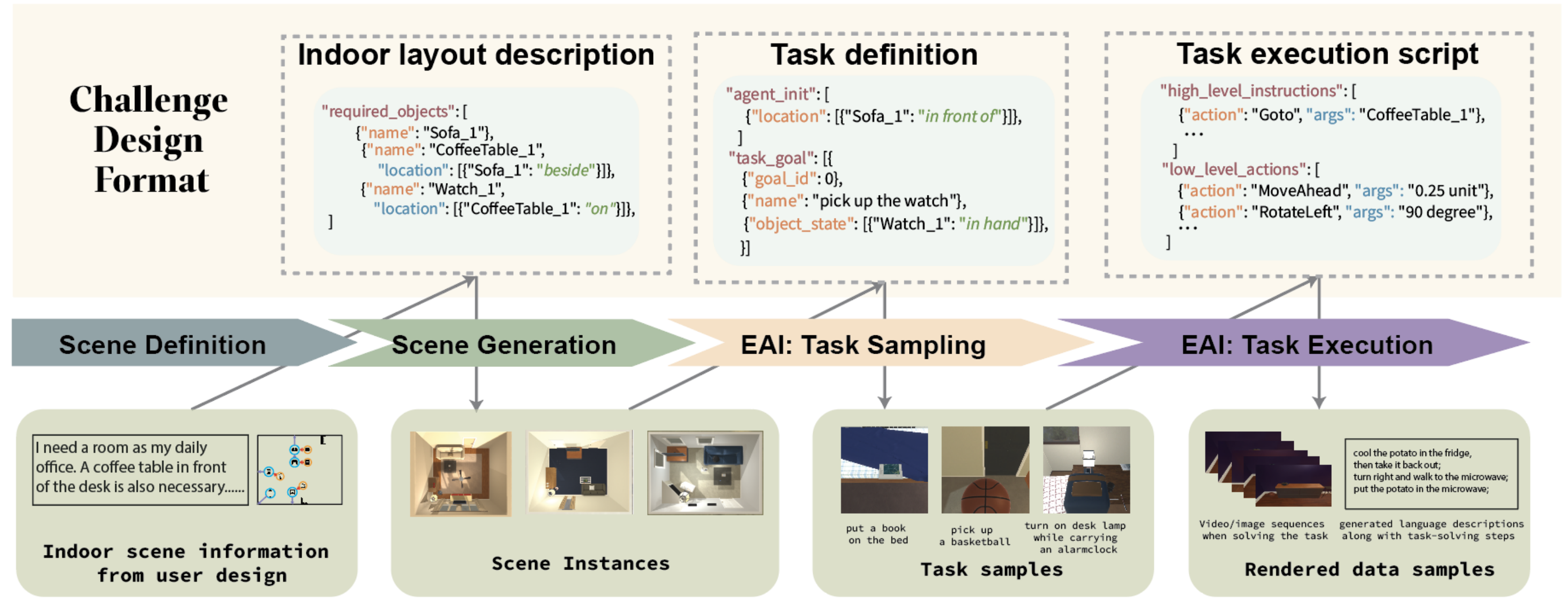}
    \caption{\textbf{The Luminous Framework.} Scene definitions constrain generated scenes, which are pragmatically evaluated via household task sampling and execution to ensure generated scene quality.}
    \label{fig:framework}
\end{figure}

\framework\ bridges the fields of indoor scene generation and EAI task completion. A well-designed indoor scene needs to support different daily tasks. Accordingly, \framework\ generates an unlimited number of randomized layouts for EAI training and evaluation, while using the task success rate of an oracle planner as an automatic metric to evaluate the quality of the generated scenes. 

\subsection{Framework Overview}
The scene generation pipeline of \framework\ consists of four stages, as shown in Figure~\ref{fig:framework}. 
First, in the \textsc{Scene Definition} stage, users specify
the required objects and, optionally, objects' relative spatial relationships.
In the \textsc{Scene Generation} stage, we propose a Constrained Stochastic Scene Generation (CSSG) algorithm to synthesize scenes whose layouts are randomized while satisfying user requirements and incorporating common sense knowledge to encourage scenes to be natural and functional. 
Next, the \textsc{Task Sampling} stage programmatically samples household tasks that are executable in the current scene. Finally, the \textsc{Task Execution} stage plans a sequence of low-level actions for the agent to execute to complete the task, and generates a series of natural language instructions to describe the agent's behavior.

\subsection{Challenge Definition Format}
We introduce the Challenge Definition Format (CDF) to concurrently support the description of indoor layouts and the execution of household tasks (Figure~\ref{fig:framework}). Learning-based indoor scene synthesis approaches are restrictive for generating EAI simulated environments~\cite{roldao20213d}. For example, these predict absolute locations for meshes, voxels, or point clouds for objects. 
By contrast, humans naturally understand the layout of an indoor scene in terms of the relative relationships
among objects, such as a coffee cup on a table, a bed against a wall, and a chair in front of a desk.
Recent scene synthesis algorithms such as Planit~\cite{wang2019planit} and 3D-SLN~\cite{luo2020end} have demonstrated the effectiveness of using a directed graph to store the relative positions of furniture. Based on this insight, we argue that relative object relationships are more important than the absolute locations of objects for understanding the functional and intrinsic utility of the room. Anecdotally, we feel specifying scene layouts through relative object relationships is more flexible and user-friendly than absolute coordinates. In the indoor layout description section of the CDF, we define the required objects that must exist in the scene, including furniture, household items, and decorations, along with the relationship among those objects, for example that a book is on a table. Figure~\ref{fig:framework} shows an example of the indoor layout description. Each entry holds the name, type, or class of an item and may optionally have its spatial relation relative to another object. In addition, similar to 3D-SLN~\cite{luo2020end}, attributes such as color, material, and size can also be attached to an entry to further describe the object.


The CDF also contains of a task definition section and a task execution script. Instead of being specified by users, these sections can be automatically generated via the task sampling stage and the task execution stage. The task definition section specifies the task to be completed within the scene. 
The execution script lists out the action sequences for completing the task.
Within the task definition section, inspired by Planning Domain Definition Language (PDDL)~\cite{haslum2019introduction, shridhar2020alfred}, the CDF defines the initial state of the scene, comprising the position of the agent and the states of objects, and the conditions for task completion, for example that a desk lamp is toggled on.
Figure~\ref{fig:framework} shows an example of an EAI task definition. 
The CDF can contain the execution script for the task in the form of human-understandable (high-level) instructions and atomic (low-level) actions.


\subsection{Constrained Stochastic Scene Generation}
\label{sec:sub:cssg}
To stochastically generate high-quality indoor scenes satisfying the layout constraints defined in the CDF, we propose a novel method: Constrained Stochastic Scene Generation (CSSG). 
Inspired by the energy-based indoor scene synthesis method~\cite{qi2018human}, 
CSSG generates scenes in a hierarchical manner, which enables great
flexibility to enforce constraints and to incorporate prior knowledge. 
First, CSSG samples the room structure, such as walls, floors, and windows, from a set of pre-defined candidates. Next, CSSG samples types, positions, and rotations of large furniture defined in the CDF. During sampling, unlike human-centric indoor scene synthesis which learns the distribution of furniture from data, CSSG generates the distribution of the position and orientation of furniture according to \textit{relationships} among furniture and room structure. Next, CSSG places objects in or on specific furniture, for example placing a coffee machine on a dining table. Finally, CSSG optionally generates decorations such as wall paintings and carpets. 

\begin{figure}
    \centering
    \hspace{-.3cm}\includegraphics[width = 1\textwidth]{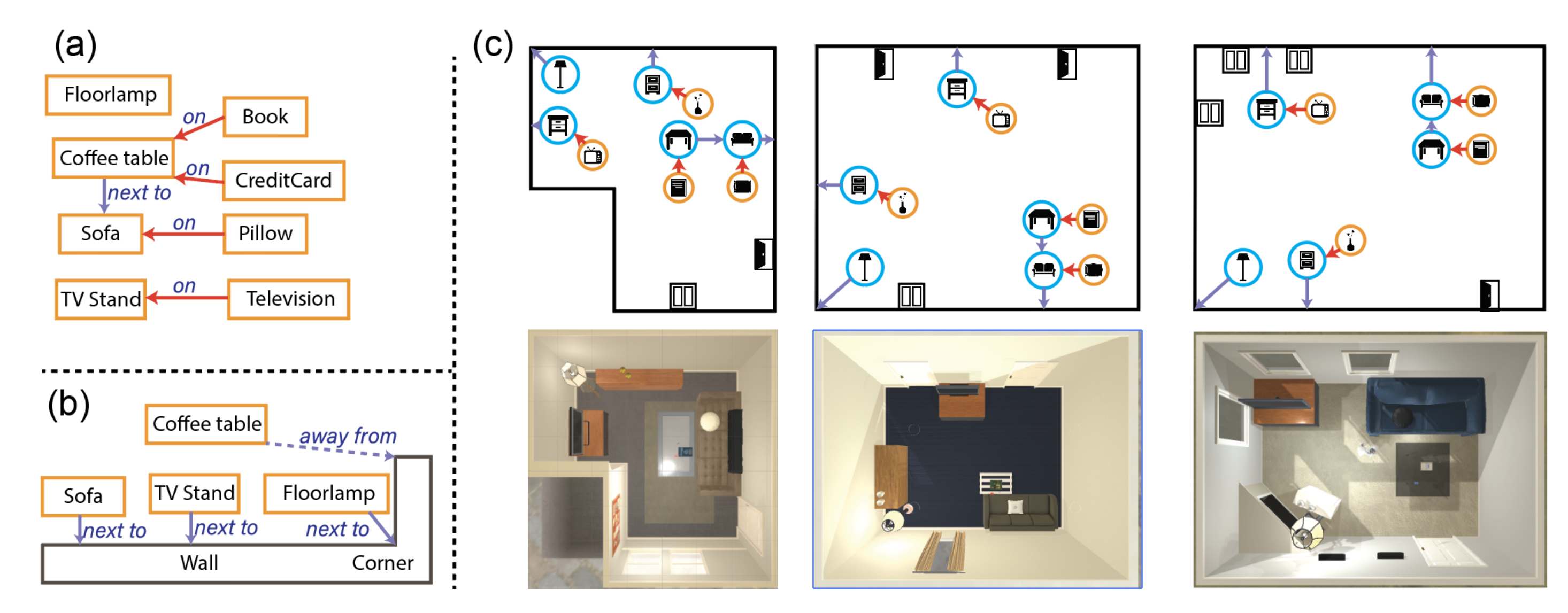}
    \caption{\textbf{CSSG illustration.} (a) explicit relationships defined in the CDF; (b) implicit relationships added by \framework; (c) sampled scenes satisfying relationships defined in (a) and (b) with different room structures.}
    \label{fig:cssg_sample}
\end{figure}

Apart from the relationships defined explicitly in the CDF file, CSSG also integrates implicit relationships based on common sense. For example, if the CDF specifies "a bed is beside a reading desk", CSSG adds an implicit rule "the bed is against the wall" when sampling the position of the bed. When multiple relationships influence the position of an object, we use a set of predefined weights for different types of relationships. Experimental results (Section \ref{sec:exp:scenequality}) show that the \textit{rule-based} CSSG with predefined weights can reasonably balance human prior knowledge with the constraints specified in the CDF thus generating meaningful and functional indoor scenes. Therefore, \framework\ adopts CSSG as the default scene generation algorithm for EAI evaluation. We refer readers to Section~\ref{sec:app:cssg} in the Appendix for details on implicit relationships, types of relationships, and predefined weights. Figure \ref{fig:cssg_sample} illustrates the scene generation pipeline of CSSG and shows several sample scenes generated by CSSG, with more in Appendix Section~\ref{sec:app:images}.

\subsection{Automatic EAI Task Sampling and Task Execution}
Another challenge of using traditional indoor scene synthesis for EAI tasks
is the lack of logic inherent to object interaction, state changes, and agent actions. It is unclear how to enable complex interaction capabilities within the framework of prior scene generation algorithms. To enable consideration of object interaction constraints, \framework\ is implemented on top of the  interactive 3D platform \thor~\cite{kolve2017ai2}, which possesses $102$ interactive object types, more than $2000$ 3D meshes, and most importantly: physical interaction mechanisms. We seamlessly connect the high-quality indoor scenes generated by CSSG and the sophisticated physical interaction logic provided by \thor. \framework\ can thus directly support many complicated EAI challenges, including but not limited to \alfred~\cite{shridhar2020alfred}, Rearrangement~\cite{batra2020rearrangement}, and RoboTHOR~\cite{RoboTHOR}. 

Given generated scenes, \framework\ can utilize the planner proposed in \alfred~\cite{shridhar2020alfred} to sample solutions to simulation tasks.
Additionally, given the tasks, \framework\ can resolve and generate appropriate scenes to support those EAI tasks. For details on task generation with \alfred, see Section~\ref{sec:lumin_alfred}. 
Note that the task generation in \framework\ does not rely on \alfred\ challenges. With the CDF used in \framework, we can easily sample an arbitrary number of simple tasks.


The task execution stage in \framework\ decomposes a household task into \textit{navigation} and \textit{interaction} tasks. \textit{Navigation} requires the agent to find an optimal route from one place to another while avoiding collisions, which is achieved by a planner inside of \framework. \textit{Interaction} often requires the agent to trigger the state change of certain object. For example, "taking a book on the coffee table" can be decomposed into the navigation part "go to a coffee table" and the interaction part "pick up the book". \framework\ applies Dijkstra's algorithm to get the shortest path for navigation, and \thor's interaction mechanism to perform the agent-object interaction.

\framework\ provides two methods to generate natural language descriptions for household tasks involving navigation and object interactions. The first method relies on a rule-based language template to generate language instructions for different tasks (See Appendix Section~\ref{sec:app:instr}). The second method uses the \textit{Speaker} model proposed in Episodic Transformer~\cite{pashevich2021episodic} that maps the low-level actions and corresponding egocentric images into generated language task instructions. 

\subsection{Accommodating Learning-based Indoor Scene Synthesis}
Apart from the energy-based approach (CSSG), \framework~incorporates two learning-based indoor scene synthesis methods, 3D-SLN~\cite{luo2020end} and Deep-synth~\cite{wang2018deep}, by training indoor-scene generators from the 3D-FRONT dataset~\cite{fu20203d}. An obstacle that hinders the application of most learning-based methods to EAI tasks are  object model discrepancies between the indoor-scene dataset and EAI simulators.
\framework\ accommodates indoor scenes generated by 3D-SLN and Deep-synth by matching model names, furniture sizes, and room shapes between 3D-FRONT and \thor, thereby providing a unified interface for learning-based approaches to train on the 3D-Front dataset and generated scenes with \thor\ assets.
For details, see Appendix Section~\ref{app:sec:iss}.

\subsection{\textsc{LUMINOUS}\ for \alfred: A Comprehensive Example}
\label{sec:lumin_alfred}
We apply \framework\ to \alfred, a benchmark for learning a mapping from natural language instructions and egocentric vision to sequences of actions for household tasks. 
The goal is to automatically generate additional data by \framework\ that shares exactly the same format as \alfred\ training and evaluation data. 

\begin{figure}
    \centering
    \includegraphics[width=0.9\textwidth]{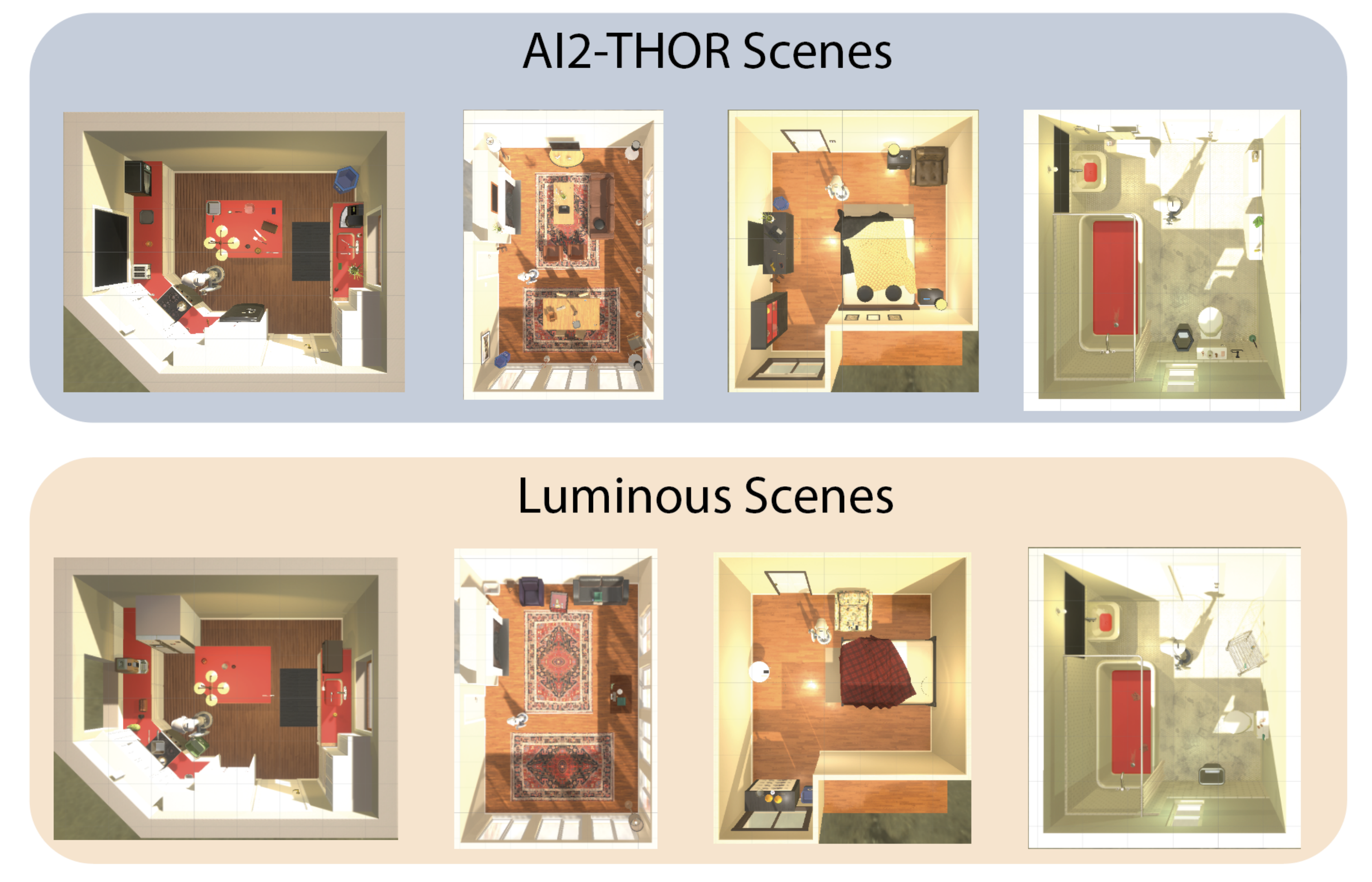}
    \caption{\textbf{Sample \thor\ and \textsc{Luminous} scenes for EAI  challenges.} For kitchens and bathrooms, \framework\ keeps more parts of the room structures.
    See Appendix \ref{sec:app:images} for more details.}
    \label{fig:images_comparison0}
    \vspace{-0.5cm}
\end{figure}

Given a trajectory $T_i$ from the \alfred\ training dataset, we employ a task parser to deduce objects and their relationships and save the scene conditions into the indoor-scene description part $I_i$ of CDF. Since each training scene in \alfred\ supports dozens of trajectories $\{T_i\}_{i=1,2,...}$, there may be some conflicting parts in their scene description $\{I_i\}_{i=1,2,...}$. For example, one task requires $\{Apple\_1\}$ to be on the countertop; another says $\{Apple\_1\}$ should be in the fridge. We propose a $merge$ operator $merge(I_1, I_2, ... ) \to \hat{I}$, where $\hat{I}$ denotes the merged links in indoor-scene description file, that tries to maximize common parts in the scene descriptions to tackle this problem. We use this merge operation for sampling indoor scene layouts $S$ by CSSG. Since \alfred\ does not change the positions of large pieces of furniture, such as fridges, sofas, and beds, the $merge$ operator records the requirements for large pieces of furniture and extracts the most common criteria for small objects (e.g., apple, cup, and book). Figure \ref{fig:images_comparison0} shows the comparison between \thor\ original scenes and \framework\ scenes generated to augment the \alfred\ challenge. 

After obtaining an indoor scene $S$, we apply two techniques to sample tasks and trajectories. The first follows the Fast-Forward Planner (FF-Planner)~\cite{shridhar2020alfred} and samples tasks and trajectories by sequentially setting initial conditions, sampling task goals, and executing trajectories. The second follows the original task design $D_i$ and directly applies the \textit{task execution} component to generate the trajectory $T'_i$. 
Locations of small objects defined by $I_i$ must be resampled for each task before execution.

The FF-Planner is slower at sampling tasks because it experiences trial and error in different sampling stages. We compare the efficiency of this method between sampling from \thor\ original scenes and from \framework-generated scenes in Section~\ref{sec:exp:scenequality}. The sampling efficiency indicates the quality of the indoor scene. The second method samples trajectories much faster since it directly applies the task design $D_i$ from original \alfred\ training data which can be quickly solved by the \textsc{Task execution} stage in \framework. We apply this method to generate a large number of scenes for the evaluation performance of different models in Section~\ref{sec:exp:eval}. 

\section{Experiments}
\label{sec:exp}
We evaluate \framework\ both quantitatively and qualitatively. Our experiments focus on answering the following questions: 1) \textit{\framework\ for indoor scene synthesis}:  Does \framework\ generate high-quality scenes that are aligned with human common sense?
2) \textit{\framework\ for EAI}: How well do the generated scenes support downstream EAI tasks? 
3) \textit{EAI task evaluation with \framework}: 
Can \framework\ generate indoor scenes that serve as reliable evaluation environments for EAI tasks? 
In addition, we discuss the insights obtained from the evaluation of state-of-the-art language-guided task completion models with larger set of unseen environments generated via \framework.

\subsection{The Quality of \textsc {Luminous}-generated Scenes}
\label{sec:exp:scenequality}
To answer the first two questions on evaluating the quality of
\framework~ generated scenes from the perspective of both 
human common sense and the capability of supporting EAI tasks, we conduct user studies and oracle task success rate.
We further demonstrate the great variety of tasks supported on scenes generated by \framework{}.

\textbf{User Studies}: 
Following the evaluation protocol proposed in~\cite{qi2018human}, we conducted user studies on Amazon Mechanical Turk comparing the quality of \bedroom\ scenes generated by \framework\ with two state-of-the-art learning-based approaches: Deep Priors~\cite{zhang2020deep} and 3D-SLN~\cite{luo2020end}.
Generated scenes are shown to users without any post-processing such as removing bad samples.
Additionally, we compared \framework\ scenes against human-designed scenes in \thor~\cite{kolve2017ai2}. 
Users were asked to evaluate scene quality, with scenes given as top-view images (Figure~\ref{fig:images_comparison0}), based on two criteria: functionality and naturalness. Functionality describes how the room layout satisfies a human’s needs for daily life.
Naturalness indicates whether the room layout is realistic.
Scales of responses range from $1$ to $5$, with 5 indicating perfect functionality or naturalness.
For every scene, we collect three ratings per metric.
The mean ratings and standard deviations are summarized in Table~\ref{tab:user_study}.
\framework~ achieves competitive performance with the human-designed scenes in \thor~\cite{kolve2017ai2}.
We ran six Welch's unpaired, two-tailed $t$-tests to compare \framework\ scores with those of \thor\ and the learning-based approaches on both metrics.
After a Bonferroni multiple-comparison correction, we find that \framework\ scenes are rated statistically significantly more functional and natural than scenes from both Deep Priors and 3D-SLN, the learning-based approaches, and not significantly differently from human-designed \thor\ scenes.


\begin{table}[t]
    \setlength{\aboverulesep}{0pt}
    \setlength{\belowrulesep}{0pt}
    \begin{small}
    \begin{tabular}{cjraarr}
    & \mcc{1}{g}{\bf Method} & \mcc{1}{c}{\bf \# Scenes,} & \mcc{1}{g}{\bf Functionality} & \mcc{1}{g}{\bf $p$-value vs.} & \mcc{1}{c}{\bf Naturalness} & \mcc{1}{c}{\bf $p$-value vs.} \\
    & & \mcc{1}{c}{\bf \# Ratings} & \mcc{1}{g}{\bf (1-5)} & \mcc{1}{g}{\bf \framework} & \mcc{1}{c}{\bf (1-5)} & \mcc{1}{c}{\bf \framework} \\
    \toprule
    \multirow{3}{*}{\bf Generated} & Deep Priors & 50, 150 & $2.40\pm1.40$ & $\pmb{\sim .0}$ & $1.78\pm1.06$ & $\pmb{\sim .0}$ \\
    & 3D-SLN & 50, 150 & $2.45\pm1.43$ & $\pmb{\sim .0}$ & $2.03\pm1.35$ & $\pmb{\sim .0}$ \\
    & \framework & 50, 150 & $\pmb{4.13}\pm1.00$ & & $\pmb{3.83}\pm1.11$ & \\
    \midrule
    \bf Human & \thor & 30, 90 & $4.23\pm0.97$ & $.416$ & $3.68\pm1.07$ & $.308$ \\
    \bottomrule
    \end{tabular}
    \end{small}
    \caption{\textbf{Human subjects' ratings of the functionality and naturalness of \bedroom\ scenes.}
    \framework\ is rated statistically significantly better than existing, state-of-the-art generation methods.}
    \label{tab:user_study}
\end{table}

\textbf{Task Success Rate:} Our proposed framework for indoor scene generation aims to promote better training and evaluation of the Embodied AI tasks.
We show that, powered by the Constrained Stochastic Scene Generation strategy, \framework{} procedurally generates indoor scenes that can produce high-quality trajectories for downstream navigation and object manipulation tasks in a comparable level of efficiency even to the manually-designed scenes provided by the \alfred\ \cite{shridhar2020alfred} dataset.
We adopt the same task sampling strategy as in the \alfred\ dataset, which roughly samples 200 tasks for each of the 7 task types (Pick \& Place, Stack \& Place, Examine in Light, etc.)
The tasks designed in the \alfred\ dataset involve long-horizon navigation and object manipulations in indoor scenes and are very challenging such that even those sampled in the hand-designed scenes fail to be solved most of the time by a carefully-tuned Planning Domain Definition Language (PDDL) rule-based \cite{aeronautiques1998pddl} motion planner.
Here we present the task success rate for a given set of scenes, defined as the percentage of tasks randomly sampled in the scenes that can be successfully solved by a rule-based, oracle planner.
To make a fair comparison, we use the same sampling strategy and motion planner provided by the \alfred\ dataset.
As similar to the training fold in \alfred, we construct 108 scenes by using \framework{} (26 scenes for each of the 4 room types).
We compare the task success rate of these scenes with the rate of the manually designed scenes from \thor~\cite{kolve2017ai2}.
Our scene generation algorithm is automatic, and does not leverage knowledge of the motion planner in \alfred\ that is tailored towards \thor\ scenes.

\textbf{Subgoal Statistics:}
Scenes generated by \framework{} support a large variety of (sub-)tasks introduced as ``subgoals'' in the \alfred\ dataset.
Each task in \alfred\ consists of several subgoals ranging from navigation to object manipulations such as ``SliceObject'' and ``ToggleObject''.
In total there are 8 types of subgoals and we calculate the statistics of these subgoals in tasks sampled from scenes as described above.
See Table \ref{tab:task_sb_completion} (Right) for the comparison between \framework{} and \thor. This subgoal level evaluation further reveals appealing properties of \framework. For example, \framework\ achieves 17\% task success rate in the GotoLocation subgoal, which indicates the generated scene has a comparable connectivity with human-created scenes in \thor\ and the robot can move freely across a large portion of scene using a simple planner that does not account for held-object collisions.



\begin{table}[t]
\begin{tabular}{cc}
    \setlength{\aboverulesep}{0pt}
    \setlength{\belowrulesep}{0pt}
    \begin{small}
    \begin{tabular}{lar}
    & \mcc{2}{c}{\bf Task Success Rate} \\
    & \thor & \framework \\
    & (Human) & (Generated) \\
    \toprule
    Pick \& Place & .33 & .13 ($\Delta$-\bad{.20}) \\
    Pick Two \& Place & .10 & .06 ($\Delta$-\good{.04}) \\
    Examine in Light & .55 & .59 ($\Delta$\phantom{-}\good{.04}) \\
    Clean \& Place & .18 & .17 ($\Delta$-\good{.01}) \\
    Heat \& Place & .19 & .09 ($\Delta$-\bad{.10}) \\
    Cool \& Place & .07 & .07 ($\Delta$\phantom{-}\good{.00}) \\
    Stack \& Place & .05 & .09 ($\Delta$\phantom{-}\good{.04}) \\
    \midrule
    Overall & .21 & .17 ($\Delta$-\good{.04}) \\
    \bottomrule
    \end{tabular}
    \end{small}
    &
    \setlength{\aboverulesep}{0pt}
    \setlength{\belowrulesep}{0pt}
    \begin{small}
    \begin{tabular}{lar}
    & \mcc{2}{c}{\bf Subgoal Success Rate} \\
    & \thor & \framework \\
    & (Human) & (Generated) \\
    \toprule
    Heat Object & .19 & .09 ($\Delta$-\bad{.10}) \\
    Cool Object & .07 & .07 ($\Delta$\phantom{-}\good{.00}) \\
    Clean Object & .18 & .17 ($\Delta$-\good{.01}) \\
    Slice Object & .11 & .09 ($\Delta$-\good{.02}) \\
    Put Object & .15 & .10 ($\Delta$-\bad{.05}) \\
    Toggle Object & .55 & .59 ($\Delta$\phantom{-}\good{.04}) \\
    Pickup Object & .22 & .18 ($\Delta$-\good{.04}) \\
    Goto Location & .21 & .17 ($\Delta$-\good{.04}) \\
    \bottomrule
    \end{tabular}
    \end{small}
\end{tabular}
\vspace{0.2cm}
\caption{
    \textbf{Left: Task Success Rate.} 
    For most task types, the loss in success rate between \thor\ human-created scenes and \framework\ generated scenes is \good{less than 5\%}, and for some tasks success rate improves.
    \textbf{Right: Subgoal Success Rate.} Multiple subgoals are carried out for each task.
    The loss in success rate in \framework\ generated scenes is usually \good{less than 5\%}, and sometimes improves.
    }
    \label{tab:task_sb_completion}
    \vspace{-0.5cm}
\end{table}

\subsection{\textsc{LUMINOUS} as an EAI Evaluation Platform}
\label{sec:exp:eval}

\begin{table}[t]
    \setlength{\aboverulesep}{0pt}
    \setlength{\belowrulesep}{0pt}
    \begin{small}
    \begin{tabular}{ljlarararar}
    & & & \mcc{8}{c}{\bf Trajectories per Task Type} \\
    \bf Split & \bf Scene & & Pick & Pick Two & Examine & Clean & Heat & Cool & Stack & \bf Overall \\
    \toprule
    \multirow{2}{*}{\splits{Seen}} & \thor & (S) & 46 & 33 & 29 & 27 & 34 & 38 & 34 & \bf 251 \\
    & \framework & (S+) & 226 & 167 & 236 & 210 & 163 & 202 & 201 & \bf 1405 \\
    \midrule
    \multirow{2}{*}{\splits{Unseen}} & \thor & (U) & 30 & 24 & 54 & 36 & 42 & 36 & 33 & \bf 255 \\
    & \framework & (U+) & 27 & 18 & 178 & 56 & 21 & 56 & 79 & \bf 435 \\
    \bottomrule
    \end{tabular}
    \end{small}
    \caption{\textbf{Validation Trajectory Counts by Task Type.} \alfred\ trajectories were sampled from both human-created \thor\ scenes and generated \framework\ scenes to evaluate EAI agents.}
    \label{tab:alfred_task_count}
    \vspace{-0.5cm}
\end{table}


We use \framework\ to provide two different settings to evaluate state-of-the-art inference models for the \alfred\ challenge. All simulated scenes, trajectories, 
and task instructions are generated by \framework. In the first setting, we use the room structures (the shape of floor, wall, and ceiling) in the 
\splits{unseen} validation set of \alfred, and then apply \framework\ to randomize the scene layouts and sample the tasks and trajectories under the same room structures. For each of the four rooms' structures in 
the validation \splits{unseen} set, we sample four room layouts and dozens of tasks. For each task, we sample one trajectory to solve the task. In total, we generate 
$16$ indoor scenes and $435$ trajectories. In the second setting, we randomly 
take 10 room structures in the \textit{training} set of \alfred\ for each room 
type (\kitchen, \livingroom, \bedroom, and \bathroom). Then, with the $40$ room structures, we randomize one layout and dozens of tasks for each. The second setting produces $1405$ trajectories for evaluating EAI models, which is an order of magnitude larger than \alfred\ \splits{unseen} in terms of both
task numbers and scene numbers. Table \ref{tab:alfred_task_count} summarizes the number of trajectories for each task type in \alfred\ validation \splits{seen}, 
\splits{unseen}, and the two evaluation settings empowered by \framework. 

With the aforementioned four test settings, we evaluate three top-ranked models for \alfred\ challenge: MOCA~\cite{singh2020moca}, Episodic Transformer (ET)~\cite{pashevich2021episodic}, and HiTUT~\cite{zhang2021hierarchical} on \framework\ validation settings. We denote the first validation setting as Unseen Plus (U+) and the second as Seen Plus (S+).
For the validation performance of MOCA and HiTUT on \alfred\ \splits{seen} and \splits{unseen}, we directly report their performance described in the paper.  For the experimental results of ET, we evaluate its performance based on the checkpoints provided by the authors of ET.  

In Table~\ref{tab:evaluation}, we show the overall performance and per-task type's for MOCA, ET, and HiTUT. First, we found that the relative performance of the three models in our setting is generally consistent with \alfred's overall generalization performance, where HiTUT achieves the best performance among the three models, and ET outperforms MOCA. It indicates that the models that perform well in the \alfred\ challenge adapt to our randomized scenarios and tasks. 
However, comparing the evaluation results in unseen environments (U vs U+), there is a notable drop in generalization performance when we increase the number of test scenes from 4 to 16. This confirms that the current evaluation in \alfred\ 
might not provide "true" generalization evaluation and highlights the significance of \framework\ for the embodied AI research.
Second, we notice that the performance under S+ is similar to \alfred\ \splits{unseen} (U) in terms of large performance drop compared to \alfred\ \splits{seen} (S),  even though the scenes and tasks generated by \framework\ share the same room structure (including walls, windows, doors, etc.) with scenes in \alfred's training. The randomized layouts from \framework\ that produce different locations of objects introduce extra difficulties for the models to accomplish tasks. It is worth noting that the high success rate of Pick tasks is due to \framework\ place the object in the edge of receptacles (e.g., table, shelf, sofa, etc.). This provides a broader range of areas for the robot to pick up the objects and thus leads to a much higher success rate than other task types. 
\begin{table}[t]
    \setlength{\aboverulesep}{0pt}
    \setlength{\belowrulesep}{0pt}
    \begin{small}
    \begin{tabular}{caarraarraarr}
    & \mcc{12}{c}{\bf \alfred\ Inference Model} \\
    & \mcc{4}{c}{\bf MOCA} & \mcc{4}{c}{\bf ET} & \mcc{4}{c}{\bf HiTUT} \\
    \bf Task & \mcc{1}{g}{S} & \mcc{1}{g}{S+} & \mcc{1}{c}{U} & \mcc{1}{c}{U+} & \mcc{1}{g}{S}  & \mcc{1}{g}{S+} & \mcc{1}{c}{U} & \mcc{1}{c}{U+} & \mcc{1}{g}{S} & \mcc{1}{g}{S+} & \mcc{1}{c}{U} & \mcc{1}{c}{U+}  \\
    \toprule
    Pick                       & .295  & .131  & .005 & .429 & .500 & .227 & .040 & .381  & .359  & .314 & .260 & .259 \\
    Cool                       & .261  & .000 & .070  & .000   & .532  & .035 & .010 & .018   & .190  & .035 & .046  & .034   \\
    Stack                      & .052 & .000  & .018  & .000    & .296  & .025 & .028 & .000   & .122  & .065 & .073  & .038   \\
    Heat                       & .158 & .000 & .027  & .000    & .458  & .000 & .074 & .000     & .140  & .061 & .119 & .000   \\
    Clean                      & .223 & .000 & .024  & .000    & .482  & .129 & .170  & .109 & .500  & .229 & .212 & .232 \\
    Examine                    & .202  & .000 & .132 & .000   & .426  & .072 & .070 & .034   & .266 & .173 & .081  & .067  \\
    Pick Two                   & .112 & .011 & .011  & .000    & .419  & .034 & .051 & .000   & .177 & .096 & .124 & .111  \\ \hline
    Average                    & .186 & .022 & .038  & .021    & .448  & .078 & .066 & .048   & .252 & .147 & .124 & .090  \\ \hline
    \end{tabular}
    \end{small}
    \caption{\textbf{Success rate on \alfred\ tasks across validation splits.}
    S: \alfred\ \splits{seen}; U: \alfred\ \splits{unseen}; U+ \splits{Unseen Plus via \framework}; S+ \textit{Seen Plus via \framework}.
    Note that all \alfred\ models, in both \splits{seen}- and \splits{unseen}-based layouts, suffer loss of performance when generalizing to generated \framework\ scenes for nearly every task.
    }
    \label{tab:evaluation}
    \vspace{-0.5cm}
\end{table}

\section{Conclusion}
\label{sec:con}


We introduced \framework, a framework to \textit{illuminate} general indoor scene generation for EAI challenges. \framework\ generates large-scale, high-quality simulated indoor scenes that are competitive with manually designed scenes in terms of naturalness and their ability to support various EAI tasks. Extensive empirical results on  language-guided task completion challenges demonstrate the effectiveness of \framework\ to serve as 
a reliable and useful EAI evaluation platform.  


\newpage 
\bibliographystyle{unsrt}
\bibliography{citations}

 \newpage 
 \appendix

\section{\framework}

\subsection{Details on incorporating Learning-based Indoor Scene Synthesis to \framework}
\label{app:sec:iss}
As we shown in Figure \ref{fig:framework},  the overall structure of \framework{}  mainly consists of three components. First, we propose a unified representation of indoor scene processing, providing various interfaces for data processing, making the original data in different formats required by different models: e.g. RGB images, bounding boxes with object types, etc. After that, different data formats are used as inputs to different models for training indoor scene generation models. It is worth noting that we unify the model-generated scene formats again, allowing us to use the same scene rendering tools to automatically visualize the scenes. Finally, we provide different testing interfaces to uniformly evaluate the quality of various algorithm-generated scenarios.

\textbf{Data processing}\\
Since our ultimate task is to provide indoor scenes as experimental environments for Embodied AI, the data we target should provide a full set of information about the indoor scenes: e.g., house structure, furniture models, and object placement information. Luminous  selects three data sources for data processing: mesh information from 3D-FRONT~\cite{fu20203d}, and game designs from \thor~\cite{kolve2017ai2}. In the data processing, we first unify the names of items in different datasets (e.g.  \textit{picture} $=$ \textit{painting}, \textit{bedside cabinet} $=$ \textit{nightstand}). The full list of unified furniture and object names are attached in the appendix. Then we normalize the coordinated w.r.t. locations and rotations. We also normalize room scales. Finally, according to different formats of the training data for different methods, we generally provides three different data formats: RGB-D images, semantic segmentation, and bounding boxes together with object types and rotations. 

\textbf{Scene Synthesis}\\
Luminous provides some state-of-the-art algorithms for indoor scene synthesis. We chose Python as programming language ,and Pytorch  for deep learning. We have carefully referred to the source code of these these methods. However, for the reason such as missing public training dataset, and the compromise we have made for unifying data formats (e.g. \textit{double bed} $\to$ \textit{bed}), the re-implemented performance in Luminous for those methods may differ from the original one.

 \subsection{Constrained Stochastic Scene Generation}
\label{sec:app:cssg}
We consider the problem of indoor scene generation under certain constraints represented by text descriptions~\cite{ma2018language} or scene graphs~\cite{luo2020end}. In our baseline, each constraint not only defines the type of an object, but also optionally describes the object's relationship with others in the scene. In detail, a constraint $c_i$ provides the information for placing object $i$ by defining its type $o_i$ (e.g. \textit{bed}), and a set of relationship with others $R_i = \{ rel(i, j_k)\}_{k=1,2,...}$, where $j_k$ stands for another object in the scene and $rel(\cdot, \cdot)$ specifies the relationship between two objects (e.g. \textit{bed} \textit{\textbf{beside}} \textit{window}).

Given a set of constraints $\{c_i\}_{i=1,2,...}$ and the room structure (the shape of floor, wall and ceiling), an indoor scene is sampled from a sequential process of three layers. The first layer samples pieces of \textbf{furniture} that represent the overall function of the room and can be placed directly on the floor, such as \textit{bed, dinning table}, and \textit{refrigerator}. The second layer samples \textbf{objects} that are usually supported by another piece furniture such as \textit{book, pen}, and \textit{coffee machine}. Finally, the third layer samples \textbf{decorations} in the scene such as \textit{painting} and \textit{carpet}.

In each layer, we empirically defined the priority value $q(i)$ as the order for placing furniture according to object types. For example, we prefer to place \textit{desk} before placing \textit{chair}: $q(desk) > q(chair)$. Besides, we limit the constraints that can be represented by a direct acyclic graph (DAG) and resolve the relationship between objects to ensure that when calculating $rel(i, j_k)$, we have $q(i) > q(j_k)$. For example, if the text description says \textit{a desk is in front of a chair}, it is resolved as \textit{a chair faces a desk}. 

When placing each object, we samples the position and rotation of the object by its explicit relationship with others $\{ rel(i, j_k))\}_{k=1,2,...}$ defined previously, and implicit relationship with others $\{ \widetilde{rel}(i, j_k))\}_{k=1,2,...}$  predefined heuristically from our prior knowledge. For example, humans are in favor of pushing the \textit{bed} up against the \textit{wall} of a \bedroom\ $(bed, (wall, against))$. 

\begin{figure}[t]
    \centering
    \includegraphics[width=0.6\textwidth]{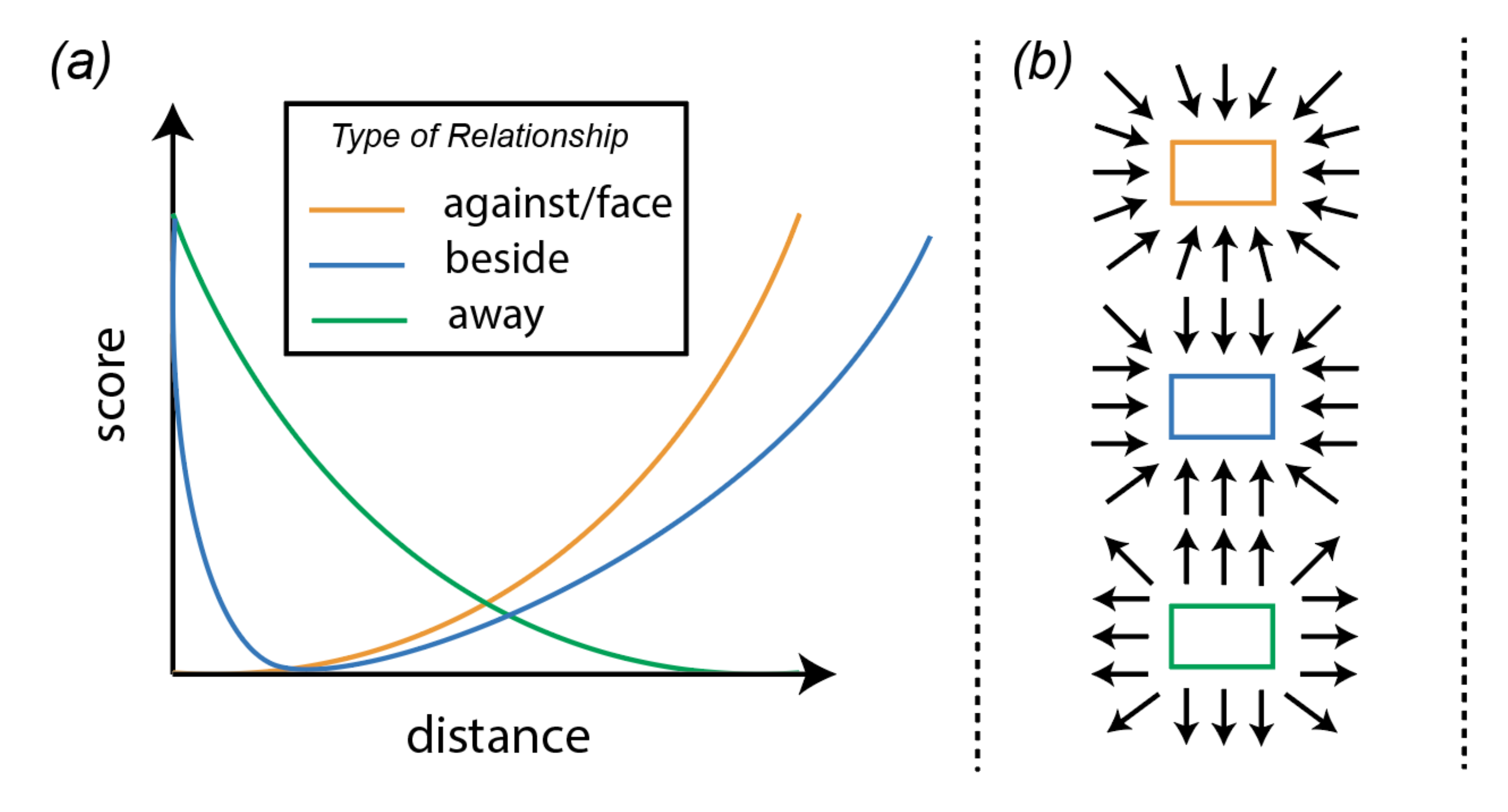}
    \caption{Illustration of how to sample the position of an object according to the type of relationship. (a) Score functions for different types of relationship, depended on the distance between the sampling position $p$ and the target object $j_k$. (b) Direction vectors suggesting the rotation $r_{p,k}$ of the object being placed on the position.}
    \label{fig:baseline_sampling}
\end{figure}

Each relationship $rel(i,j_k)$ generates a vector field in space: each position $p$ is characterized by $(s_{p,k}, r_{p,k})$, where $s_{p,k}$ is the score of the point. $s_i$ depends on the distance $d_i$ between $p$ and the target object $j_k$. Figure \ref{fig:baseline_sampling}(a) shows different types of relationship and the scores deduced by the relative distance. $r_{p,k}$, suggesting the relative rotation of placing the object, depends on the direction vector from $p$ to its target $j_k$ the type of relationship. Combining $s_{p,k}$ with parameter $w_{type(rel(i,j_{k}))}$ related only the type of relationship, we sample the position to place object $i$ according to weighed sum of scores among all relationship, and the rotation of the object at position $p$ is defined by the type of relationship which has the largest weight.
\begin{align}
    s_p &= \sum_{k} w_{type(rel(i,j_{k}))}\\
    P(p | R_i) &\propto \exp(-s_p) \\
    r_{p} &= r_{p,k'}   ~~~~k' = \arg\max \{w_{type(rel(i,j_{k}))}\}
\end{align}

\subsection{Comparison between CSSG and advanced indoor scene generation algorithms}
In Table~\ref{tab:cssg}, we summarize the properties of CSSG and other indoor scene algorithms. As the table shown, the state-of-the-art scene generation algorithms use SUNCG dataset~\cite{song2017semantic} as training , is not currently not available. It is hard to reproduce the results from those approaches. In \framework, we reproduce the learning based approaches such as 3D-SLN~\cite{luo2020end} using publicly available dataset (3D-FRONT~\cite{fu20203d}) for training. We believe this could serve as first step to provide a unified benchmark for comparing indoor scene generation algorithms.  

\begin{table}[h!]
\begin{tabular}{ccccc}
\hline
\textbf{Algorithm}                                             & \textbf{\begin{tabular}[c]{@{}c@{}}Scene graph \\ Inference?\end{tabular}} & \textbf{Constrained?}                                             & \textbf{\begin{tabular}[c]{@{}c@{}}RGBD \\ rendering?\end{tabular}} & \textbf{Dataset?}                                                   \\ \hline
PlanIT (2019)                                                  & \cellcolor[HTML]{FFFFFF}{\color[HTML]{222222} \textbf{\checkmark}}              & \cellcolor[HTML]{FFFFFF}{\color[HTML]{222222} \textbf{\checkmark}} & \cellcolor[HTML]{FFFFFF}{\color[HTML]{222222} \textbf{\checkmark}}   & \cellcolor[HTML]{FFFFFF}{\color[HTML]{222222} \textbf{unavailable}} \\
Grains (2018)                                                  & \cellcolor[HTML]{FFFFFF}{\color[HTML]{222222} \textbf{N/A}}                    & \cellcolor[HTML]{FFFFFF}{\color[HTML]{222222} \textbf{N/A}}       & \cellcolor[HTML]{FFFFFF}{\color[HTML]{222222} \textbf{\checkmark}}   & \cellcolor[HTML]{FFFFFF}{\color[HTML]{222222} \textbf{unavailable}} \\
3D-SLN (2020)                                                  & \cellcolor[HTML]{FFFFFF}{\color[HTML]{222222} \textbf{N/A}}                    & \cellcolor[HTML]{FFFFFF}{\color[HTML]{222222} \textbf{\checkmark}} & \cellcolor[HTML]{FFFFFF}{\color[HTML]{222222} \textbf{\checkmark}}   & \cellcolor[HTML]{FFFFFF}{\color[HTML]{222222} \textbf{unavailable}}            \\
\begin{tabular}[c]{@{}c@{}}Human-centric (2019)\end{tabular} & \cellcolor[HTML]{FFFFFF}{\color[HTML]{222222} \textbf{N/A}}     & \cellcolor[HTML]{FFFFFF}{\color[HTML]{222222} \textbf{N/A}}       & \cellcolor[HTML]{FFFFFF}{\color[HTML]{222222} \textbf{\checkmark}}   & \cellcolor[HTML]{FFFFFF}{\color[HTML]{222222} \textbf{unavailable}} \\ \hline
\rowcolor[HTML]{FFFFFF} 
\cellcolor[HTML]{FFFFFF}Luminous CSSG                          & {\color[HTML]{222222} \textbf{\checkmark}}                                              & {\color[HTML]{222222} \textbf{\checkmark}}                         & {\color[HTML]{222222} \textbf{\checkmark}}                           & {\color[HTML]{222222} \textbf{N/A}}                                 \\ \hline
\end{tabular}
\newline
\caption{Comparison of CSSG and state-of-the-art indoor scene generation algorithms. Scene graph inference refers to the algorithm's ability to infer the latent scene graph of the indoor scene. Some of the algorithms support taking scene graphs as constraints. The dataset for training the indoor scene synthesis model is missing due to legal issues.}
\label{tab:cssg}
\end{table}

\subsection{Implicit relationships between furniture}
We list the implicit relationships when sampling the position of the furniture. Basically, the relationships can categorizes into two types: furniture v.s. room structure, and furniture v.s. furniture. 
\begin{itemize}
    \item furniture v.s. room structure: (CounterTop, against, wall border), (TVStand, against, wall border), (Sofa, against, wall border), (border, against, wall border), (Bed, against, wall border), (Dresser, against, wall border),(Desk, against, wall border),(SideTable, against, wall border),(FloorLamp, against, wall corner), (DiningTable, away from, wall border)
    \item furniture v.s. furniture: (Chair, face, Desk), (Stool, face, DiningTable), (CoffeeTable, beside, Sofa), (DiningTable, away from, Sofa)
\end{itemize}
If multiple relationships influence the distribution of the sampling position of an object, we give the weight coefficient as $2.0$ if the relationship is from \textit{furniture v.s. room structure}, and as $1.0$ if the relationship is from \textit{furniture v.s. furniture}.




%
 \section{Task Instructions Generation}
\label{sec:app:instr}
\begin{table}[t]
\centering
\begin{tabular}{cc}
\bf High-level action & \bf Instruction candidates \\
\toprule
GotoLocation      & go to, find, walk to   \\ 
PickupObject      & pick up, take, carry   \\ 
PutObject         & put, place             \\ 
SliceObject       & slice, cut             \\ 
CoolObject        & chill, cool            \\ 
HeatObject        & heat, cook             \\ 
CleanObject       & clean, wash, rinse     \\ 
ToggleObject      & turn on                \\ 
\bottomrule
\end{tabular}
\vspace{0.1in}
\caption{Language template: mapping high-level actions to language instructions}
\label{tab:task:instr}
\end{table}

Unlike ALFRED, \framework\ obtains the natural language as high-level instructions from an automatic pipeline instead of human annotations. 

We design a language template to generate natural language instructions corresponding to the high-level instructions in ALFRED. Table \ref{tab:task:instr} shows mappings from high-level action to language instructions. The natural language instruction is generated as: 
$$[instruction~candidate] + [object~name]+ [attribute]$$
 Where the attribute specifies the receptacle for \textit{PickupObject} (e.g., pick up an apple \textit{in the fridge}), or the target location for \textit{PutObject} (e.g., put a book \textit{on the table}).

However, the language instruction for navigation can be too simple and vague if we just say \textit{go to} some place. We apply the \textit{Speaker} provided by ET to generate task instructions, especially for the navigation part. The training data come from the ALFRED dataset. The input of the \textit{Speaker} is the low -level action sequence (e.g. \textit{MoveAhead}, \textit{MoveAhead},\textit{RotateLeft}) and images from the egocentric view the agent, and the output is a piece of natural language instruction.
$$
(low~level~actions, images)\xrightarrow[\text{Speaker}]{} (language ~instructions)
$$
We refer readers to ET~\cite{pashevich2021episodic} for model details and put the generated examples in Appendix \ref{sec:app:images}
 \section{Illustration of ALFRED and \framework}
\label{sec:app:images}
In this part, we illustrate the details when we apply \framework\ for ALFRED challenge.

\subsection{Task parser}
The task parser is applied to deduce the indoor scene description $I_i$ for an ALFRED trajectory $T_i$. Specifically, the task parser would go through the low-level actions in $T_i$, and
\begin{itemize}
    \item extract the $action~args$ as required objects from actions including \textit{GotoLocaiton}, \textit{PickupObject}, \textit{ToggleObjectOn}, and~\textit{OpenObject}. For example, if the $action~args$ of \textit{GotoLocaiton} is \textit{DiningTable}, the task parser put \textit{DiningTable} into the list.
    \item extract the $action~args$ of \textit{PickupObject} as scene constraints. For example, picking up an apple on the fridge means that initially \textit{Apple} is in the \textit{Fridge}.
\end{itemize}

\subsection{Indoor scene sampling}
\begin{figure}
    \centering
    \includegraphics[width=0.85\textwidth]{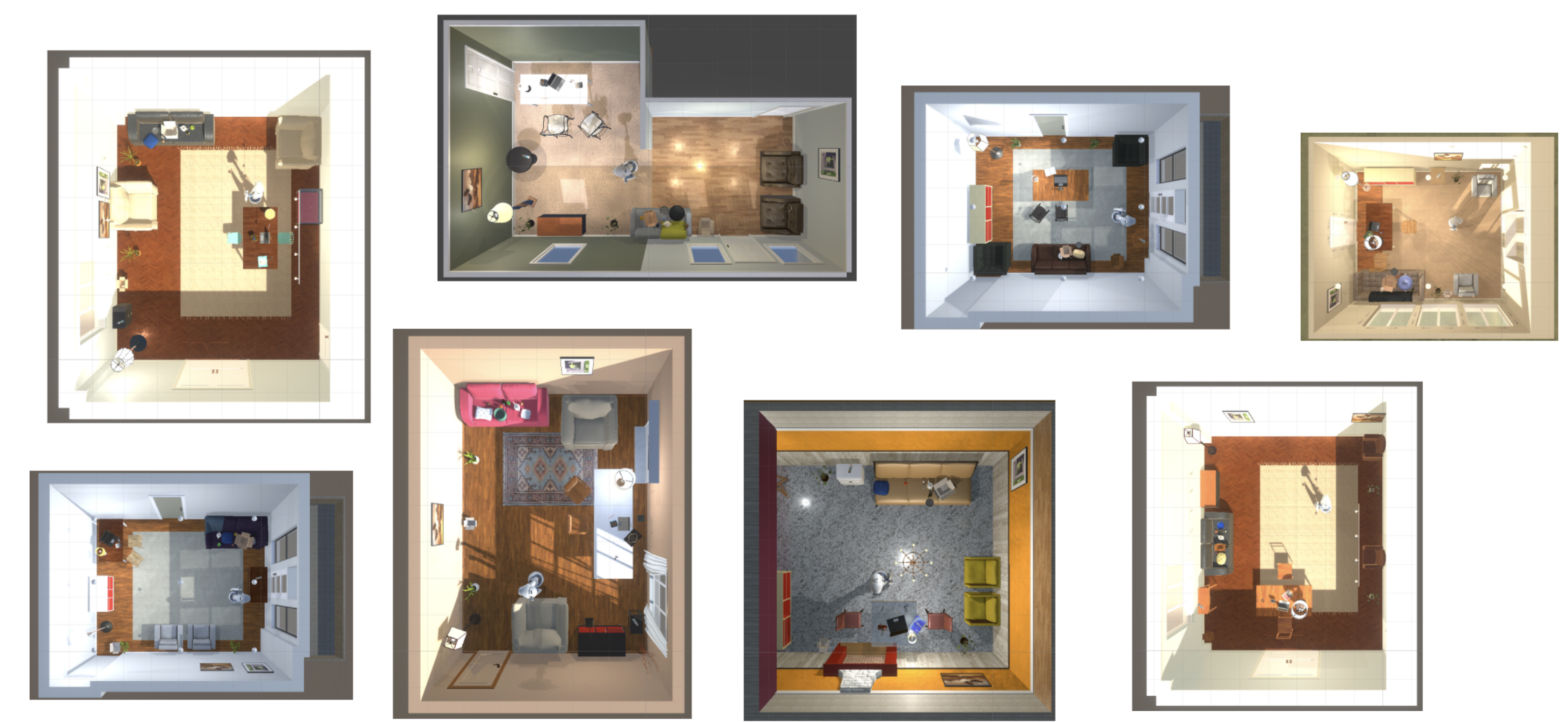}
    \caption{Living rooms sampled by \framework}
    \label{fig:lumi_livingroom}
\end{figure}

\begin{figure}
    \centering
    \includegraphics[width=0.85\textwidth]{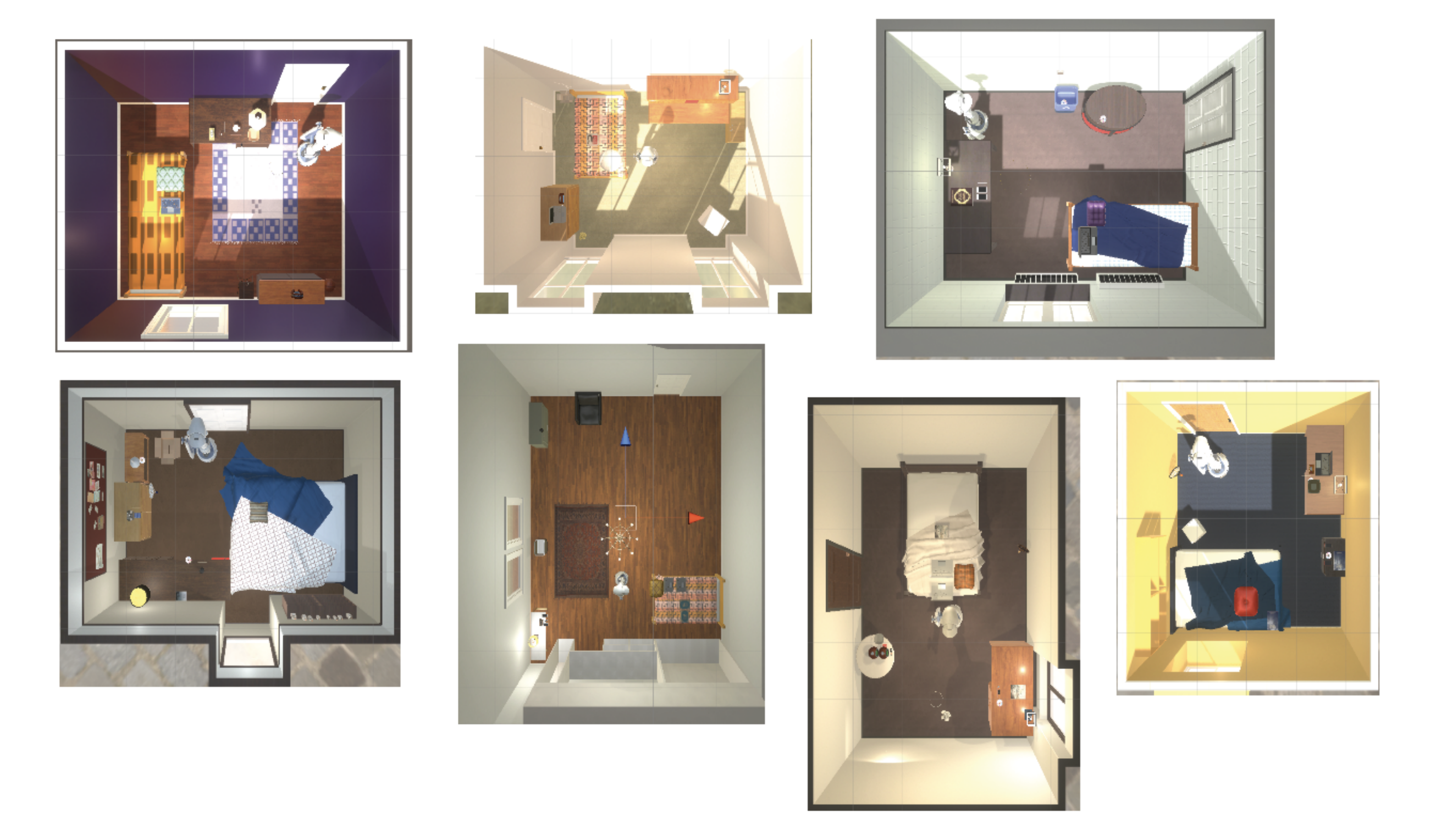}
    \caption{Bedrooms sampled by \framework}
    \label{fig:lumi_bedroom}
\end{figure}

\begin{figure}
    \centering
    \includegraphics[width=0.85\textwidth]{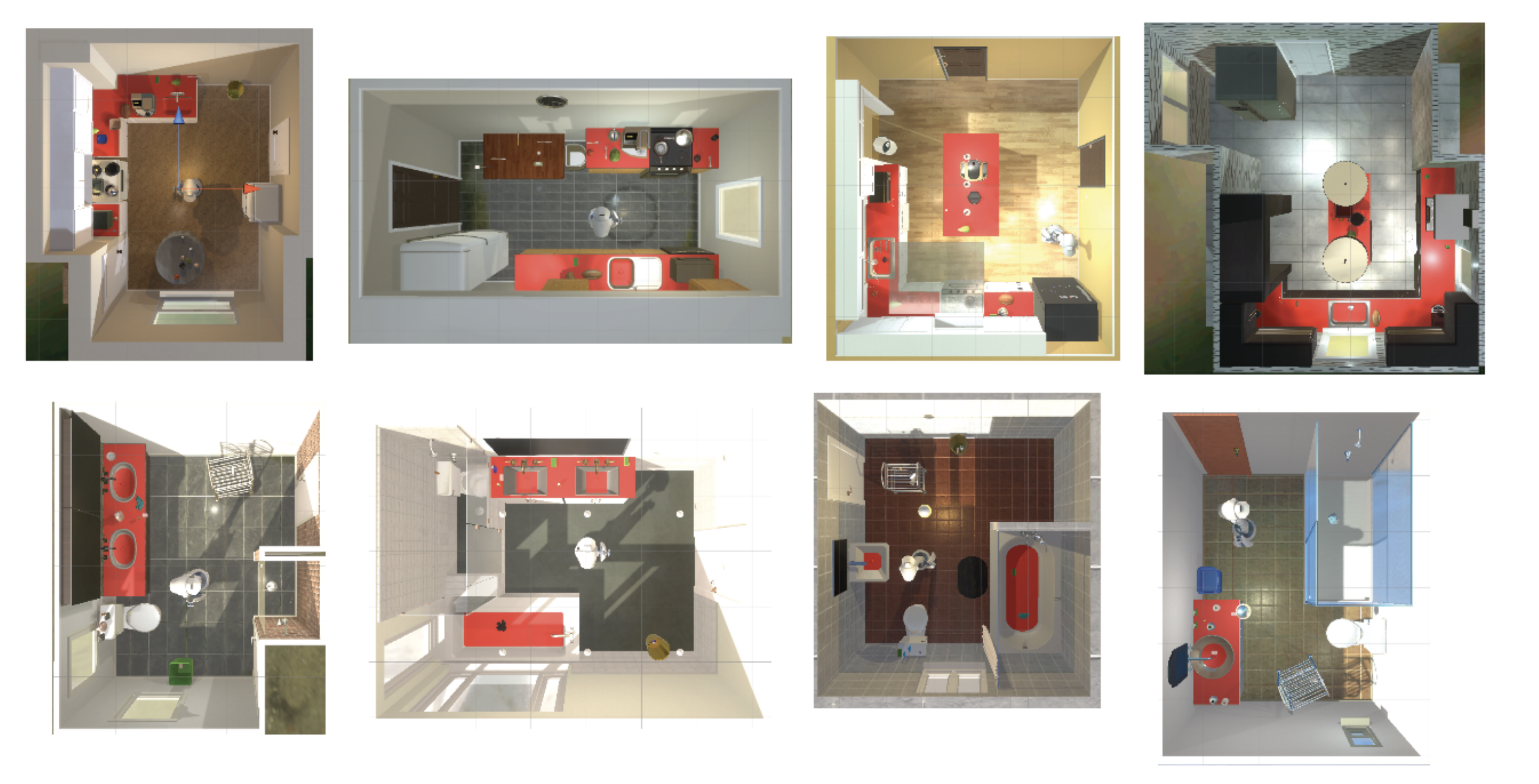}
    \caption{Kitchens and bathrooms sampled by \framework}
    \label{fig:lumi_bathroom_kitchen}
\end{figure}

For room structures of living rooms and bedrooms, \framework\ only keep \textit{wall}, \textit{ceiling}, \textit{floor}, \textit{window} and \textit{door}. For room structures of kitchens and bathrooms, \framework\ further keeps \textit{CounterTop}, \textit{Sink}, \textit{Cabinet}, and \textit{Oven}, and \textit{Bathtub}. Figure \ref{fig:lumi_livingroom}, \ref{fig:lumi_bedroom}, and \ref{fig:lumi_bathroom_kitchen} plot the scenes of different room types sampled by \framework. 

\subsection{ALFRED trajectories v.s. \framework\ trajectories}
\begin{figure}
    \centering
    \includegraphics[width=0.9\textwidth]{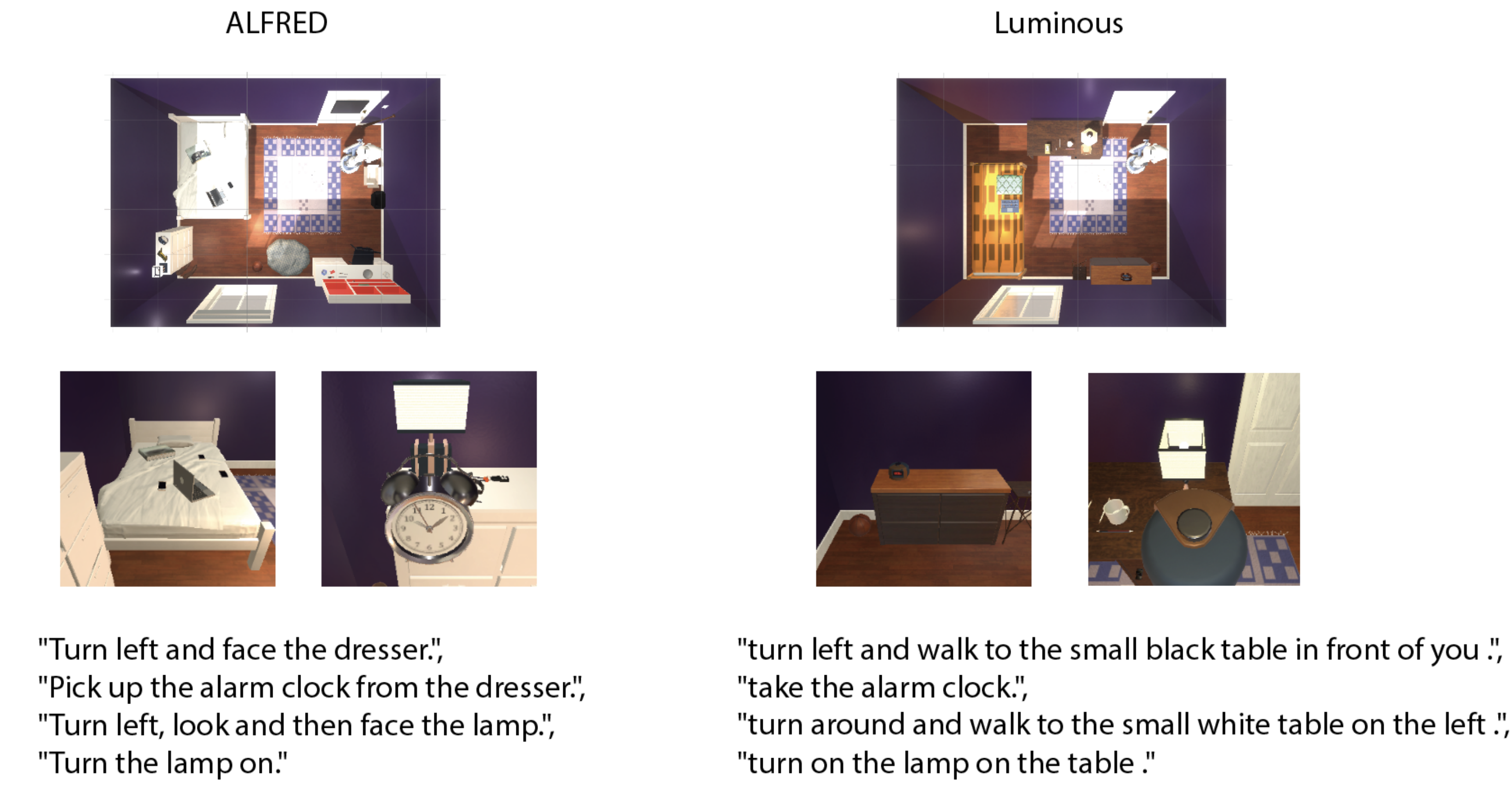}
    \caption{Comparison between ALFRED and \framework\ generated trajectories. Task name: look\_at\_obj\_in\_light-AlarmClock-None-DeskLamp; scene name: FloorPlan301\_physics; trial id:trial\_T20190907\_174127\_043461.}
    \label{fig:images_comparison1}
\end{figure}

\begin{figure}
    \centering
    \includegraphics[width=0.9\textwidth]{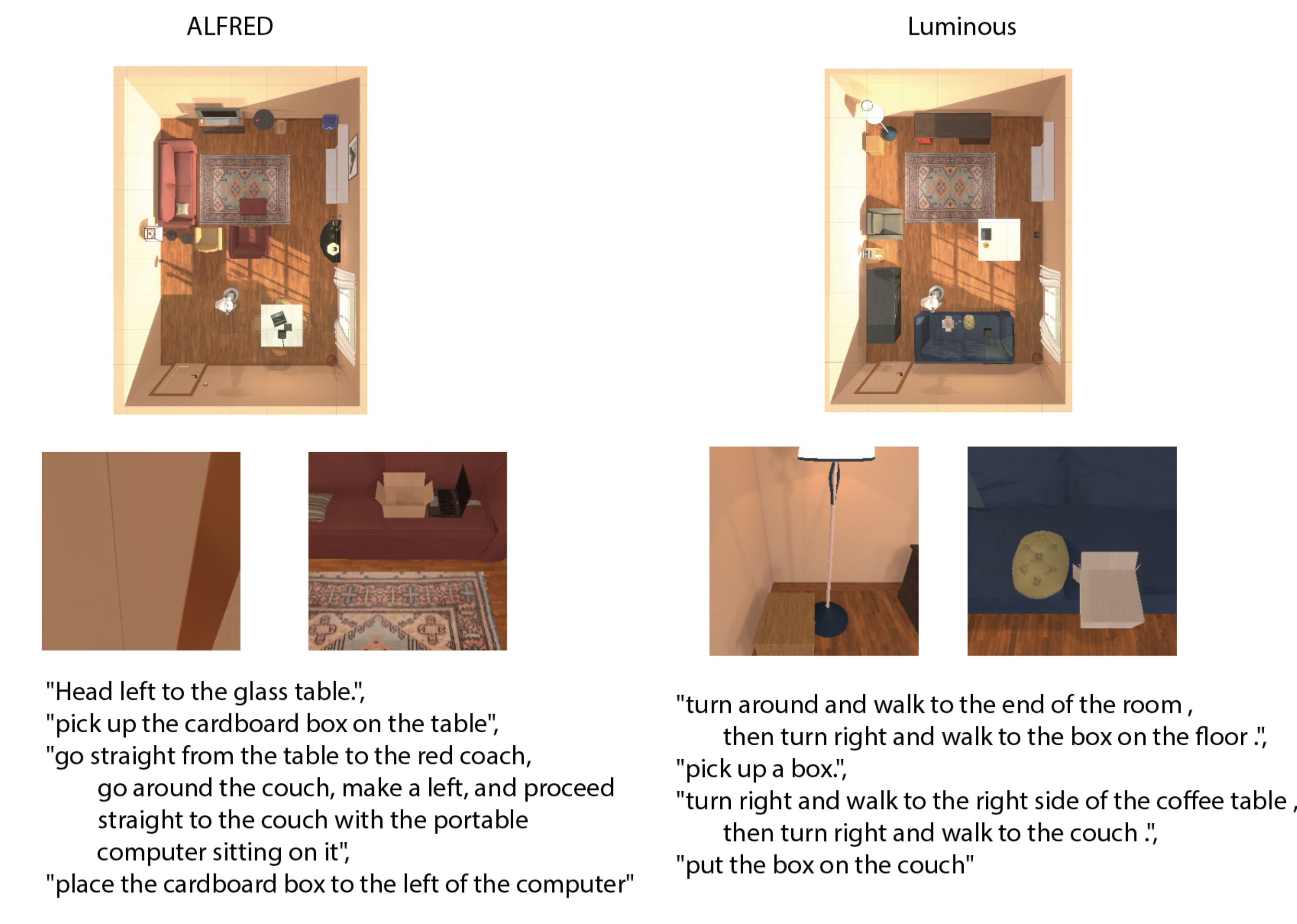}
    \caption{Comparison between ALFRED and \framework\ generated trajectories. Task name: pick\_and\_place\_simple-Box-None-Sofa-205; scene name: FloorPlan205\_physics; trial id:trial\_T20190907\_214755\_478301.}
    \label{fig:images_comparison2}
\end{figure}

We performance side by side comparison between ALFRED trajectories and \framework\ trajectories in Figure \ref{fig:images_comparison1} and \ref{fig:images_comparison2}. We plot the scene layouts, initial camera images, images after task completion and language instructions for both.

 \subsection{Hard task analysis: Heat \& Place / Cool \& Place}
 
\begin{figure}
    \centering
    \includegraphics[width=0.95\textwidth]{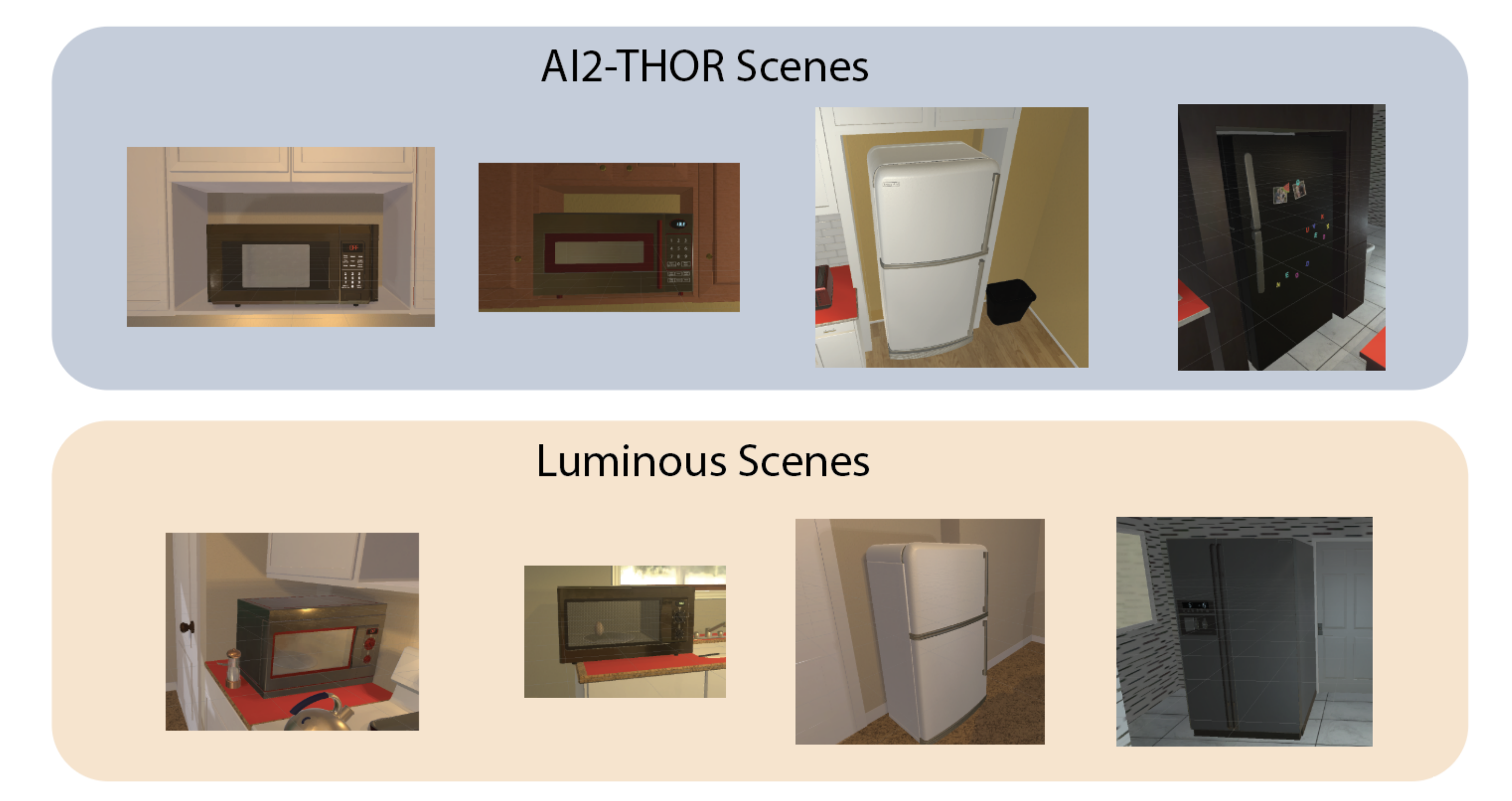}
    \caption{Different locations of the microwave and fridge in \thor\ scenes and \framework\ scenes. In AI2THOR, most microwaves and fridges are embedded in the structure of the room; in \framework, microwaves are preferred to be placed on a countertop and fridges most likely locates in a relatively open area. Such difference brings different visual experience to EAI agents.}
    \label{fig:heat_cool}
\end{figure}

We notice the low success rate for two types of tasks:  \textit{Heat \& Place} and \textit{Cool \& Place} in \framework\ scenes. The \textit{Cool} operation requires a fridge and the \textit{Heat} operation needs a microwave. We compare the layout w.r.t. the fridge and microwave between \thor\ scenes and \framework\ scenes, and we find a somewhat different set-up for them. Figure \ref{fig:heat_cool} compares the locations of the fridge and microwave. Since \thor\ scenes are manually designed.
\begin{itemize}
    \item In the task sampling stage (Table 2), the FF-Planner samples task trajectories from ground-truth knowledge of the environment and would not be influenced by visual discrepancies between ALFRED and \framework. 
    \item In the EAI evaluation stage (Table 4), the EAI agent takes the input as RGB images and images look visually different between manually designed scenes and synthesized scenes, making the agent harder to complete heat and cool tasks.
\end{itemize}


\begin{table}[t!]
\begin{tabular}{ccccc}
\hline
\textbf{Challenge}                                                                  & \textbf{Navigation?} & \textbf{Interaction?} & \textbf{\begin{tabular}[c]{@{}c@{}}Language \\ understanding?\end{tabular}} & \textbf{\begin{tabular}[c]{@{}c@{}}affordance/physics \\ understanding?\end{tabular}} \\ \hline
\begin{tabular}[c]{@{}c@{}}ObjectNav \\ (habitat, ai2thor)\end{tabular}              & YES                  & NO                    & NO                                                                        & NO                                                                                    \\
\begin{tabular}[c]{@{}c@{}}Multi-On/Rearrangement \\ (habitat, ai2thor)\end{tabular} & YES                  & PART OF               & NO                                                                        & YES                                                                               \\
InteractiveNav \\ (iGibson)                                                             & YES                  & YES                   & NO                                                                        & YES                                                                                   \\ \hline
\rowcolor[HTML]{FFFFFF} 
ALFRED \\ (ai2thor)                                                                     & YES                  & YES                   & YES                                                                       & YES                                                                                   \\ \hline
\end{tabular}
\newline
\caption{Comparison between ALFRED with other EAI tasks. Different simulators may have different requirements to EAI agents including navigation (to navigate an agent from one place to another), interaction (to interact with an object in the environment), language understanding (to follow language instructions from users), and affordance or physics understanding (to gain some knowledge for the affordance map in the scene).}
\end{table}
\section{Large-Scale Evaluation Experiments}
In this section, we conduct an additional large-scale evaluation with respect to the number of scenes. We generated 216 scenes with the same room structure as training scenes in ALFRED (including walls, floor, and windows) but randomized layouts and objects as the evaluation environments for ALFRED-like tasks. We summarize the statistics of our
evaluation datasets and performance of three state-of-the-arts in Table~\ref{tab:evaluation2}. The second column presents the number of unique configurations (including room layouts, small object locations) 
of tasks in each task type.  The third column shows the number of unique scenes/layouts (same room layout with
different small object locations count as the same scene). Comparing the results in 
Table~\ref{tab:evaluation} and Table~\ref{tab:evaluation2}, the success rate in $S+$ column evaluated 
by 40 scenes and 216 scenes maintain the similar relative performance.
Based on the above observation, we further strengthen our conclusions obtained in Section~\ref{sec:exp:eval} that
\framework\ can provide more robust and consistent evaluation results.  

\begin{table}[h!]
    \centering
    \begin{tabular}{cccararar}
    & & & \mcc{6}{c}{\bf \alfred\ Inference Model} \\
    & & & \mcc{2}{c}{\bf MOCA} & \mcc{2}{c}{\bf ET} & \mcc{2}{c}{\bf HiTUT} \\
    \bf Task  &\bf \# Trajs &\bf \# Scenes &  \mcc{1}{c}{S} & \mcc{1}{c}{S+}  & \mcc{1}{c}{S} & \mcc{1}{c}{S+}  & \mcc{1}{c}{S} & \mcc{1}{c}{S+}  \\
    \toprule
    Pick  & 1124     & 192  & .295  &  .139  & .500 & .205  & .359 & .296 \\
    Cool  & 885      & 44   & .261  & .000   & .532 & .009  & .190  & .043   \\
    Stack & 1002    & 126   & .052  & .002   & .296 & .028  & .122  & .058  \\
    Heat  & 786      & 54   & .158  & .000   & .458 & .005  & .140 & .061  \\
    Clean  & 923     & 98   & .223  & .000   & .482 & .109 & .500 &  .232\\
    Examine & 1263   & 84   & .202  &  .000  & .426 & .056  & .266  & .124   \\
    Pick Two & 944  & 168   & .112  &  .013  & .419 & .034  & .177 &  .097 \\ \hline
    Overall & 7074  &  -    & .186  &  .025  & .448 & .068 & .252 &  .137  \\ \hline
    \end{tabular}
    \newline
    \caption{\textbf{Success rate on \alfred\ tasks.} \# Trajs: number of unique task configurations; 
    \# Scenes: number of unique scene layouts in each task type; 
    S: \alfred\ \splits{seen}; S+ \splits{Seen Plus via \framework}.}
    \label{tab:evaluation2}
\end{table}

\clearpage
\section{Dataset Examples}
\begin{figure}[h!]
\vspace{-0.5cm}
    \centering
\begin{tabular}{c}
    \includegraphics[width=0.95\textwidth]{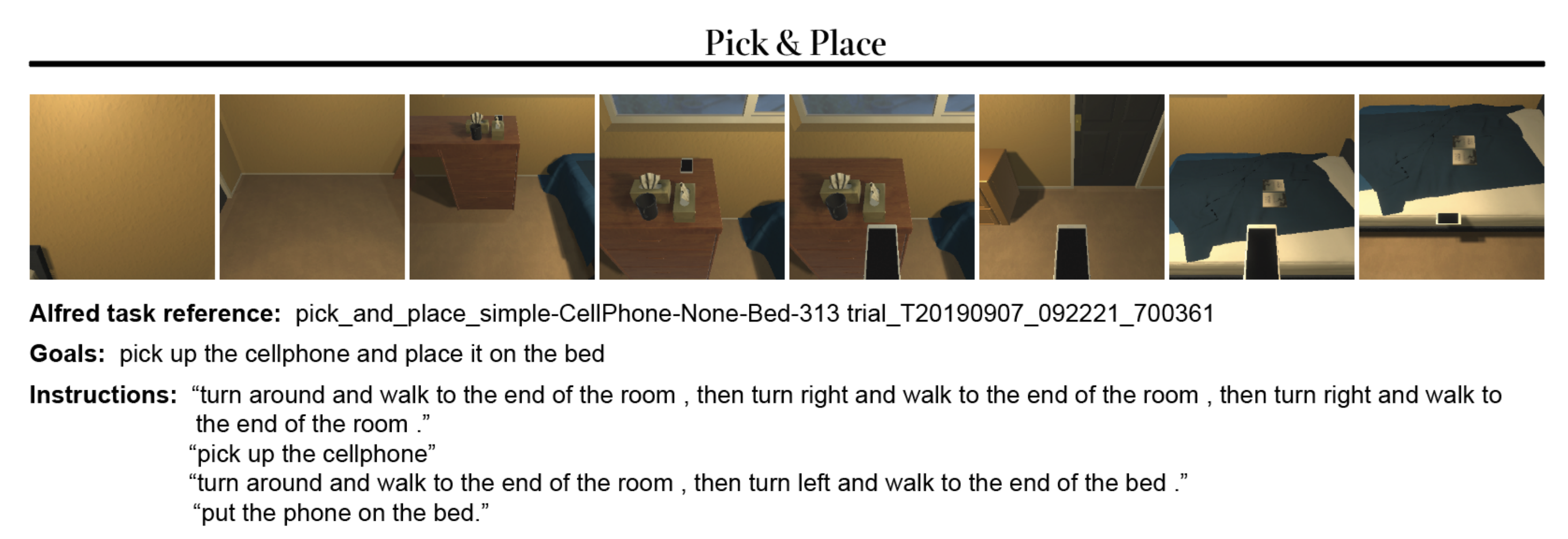}\\
    \includegraphics[width=0.95\textwidth]{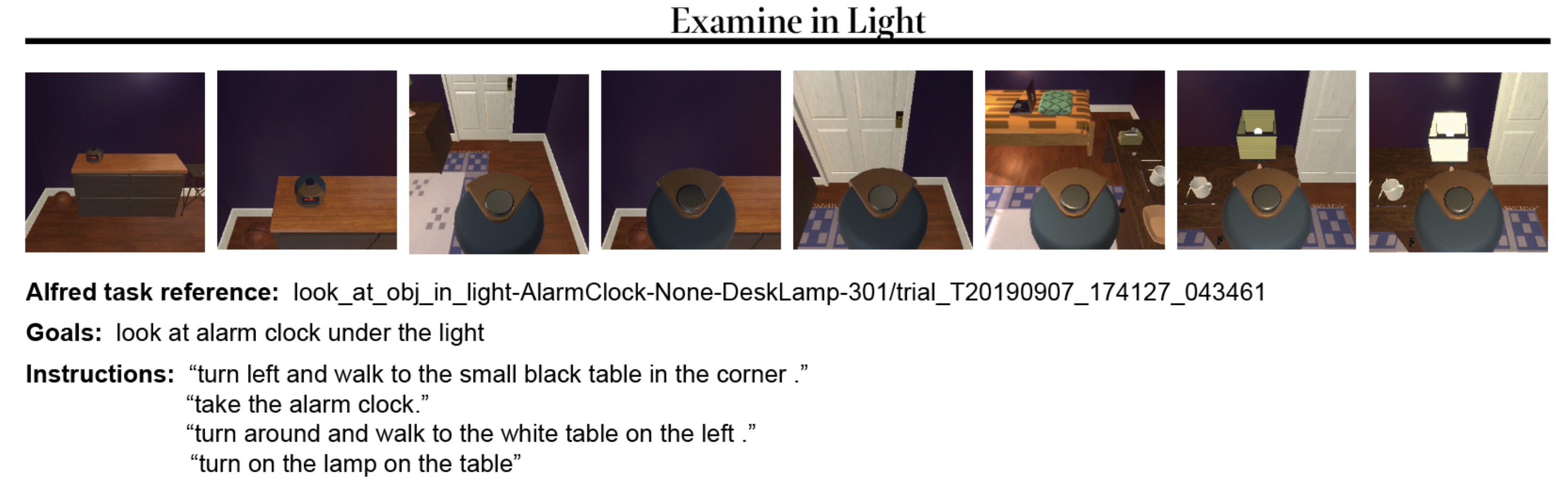}\\
    \includegraphics[width=0.95\textwidth]{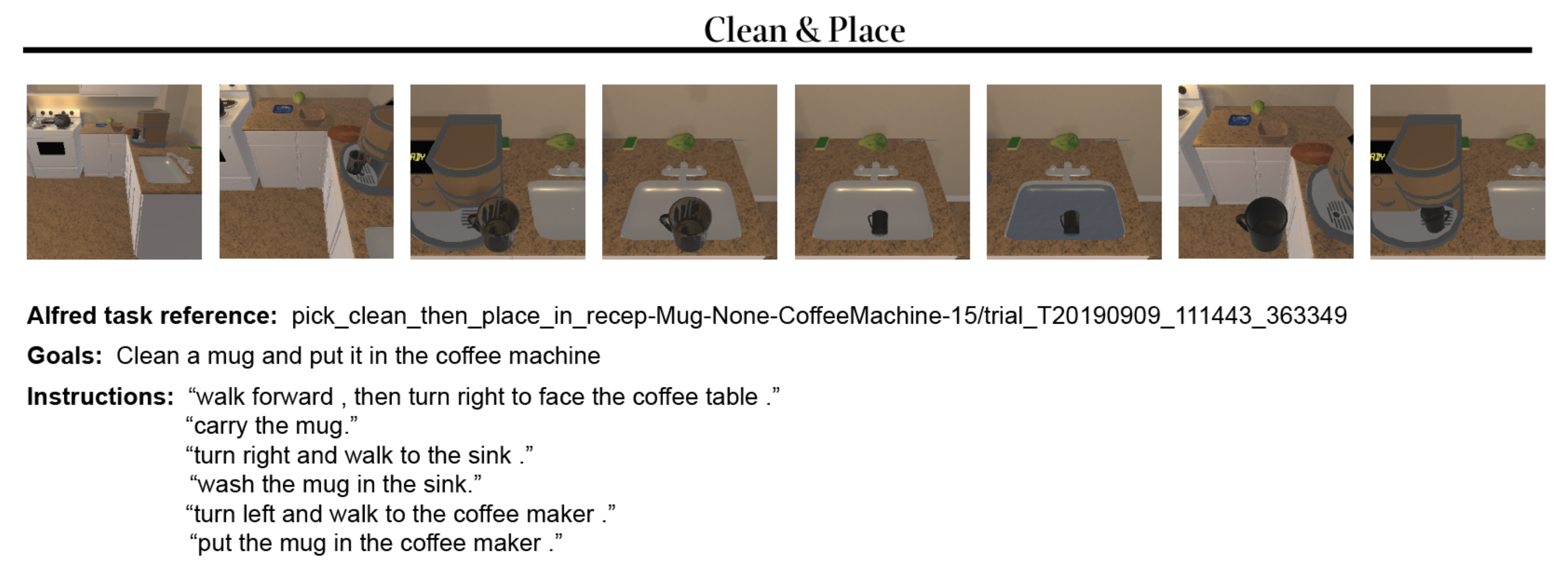}\\
    \includegraphics[width=0.95\textwidth]{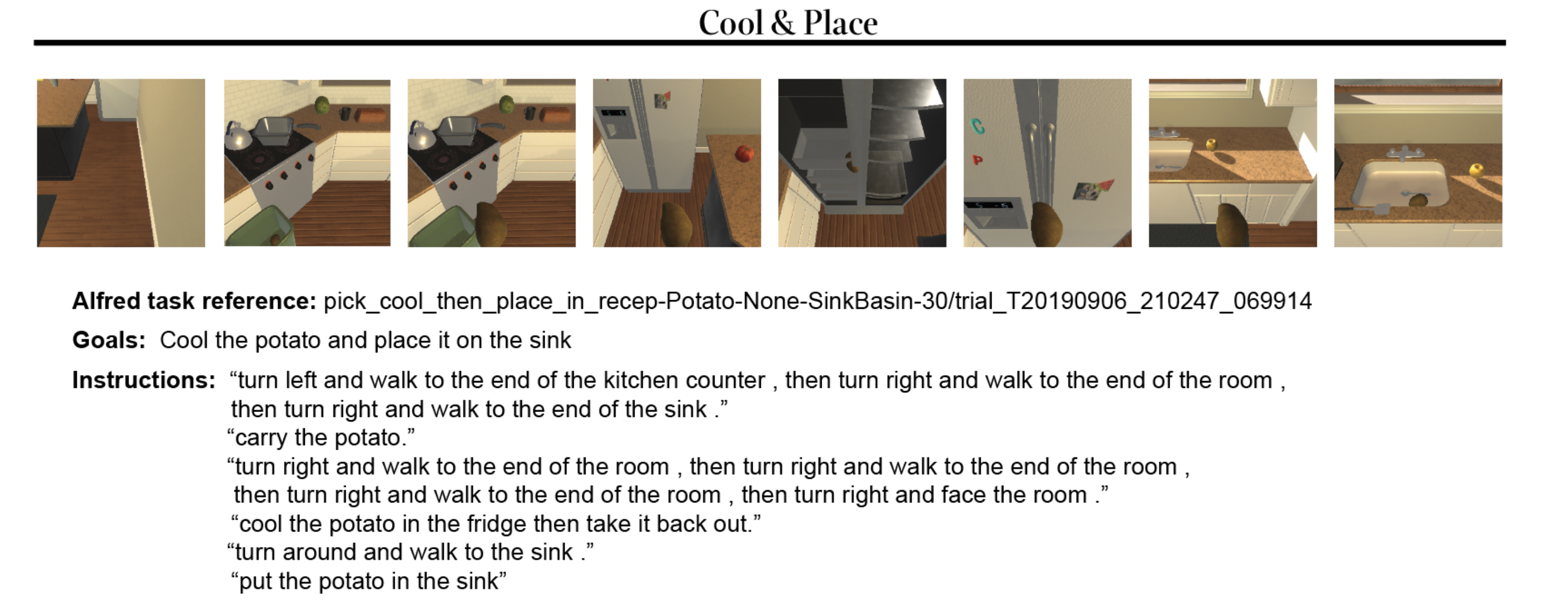}
\end{tabular}
    \caption{Dataset Examples. Automatically generated scenes, low-level actions, and language instructions.}
    \label{fig:my_label1}
\end{figure}

\begin{figure}[h!]
    \centering
\begin{tabular}{c}
    \includegraphics[width=0.95\textwidth]{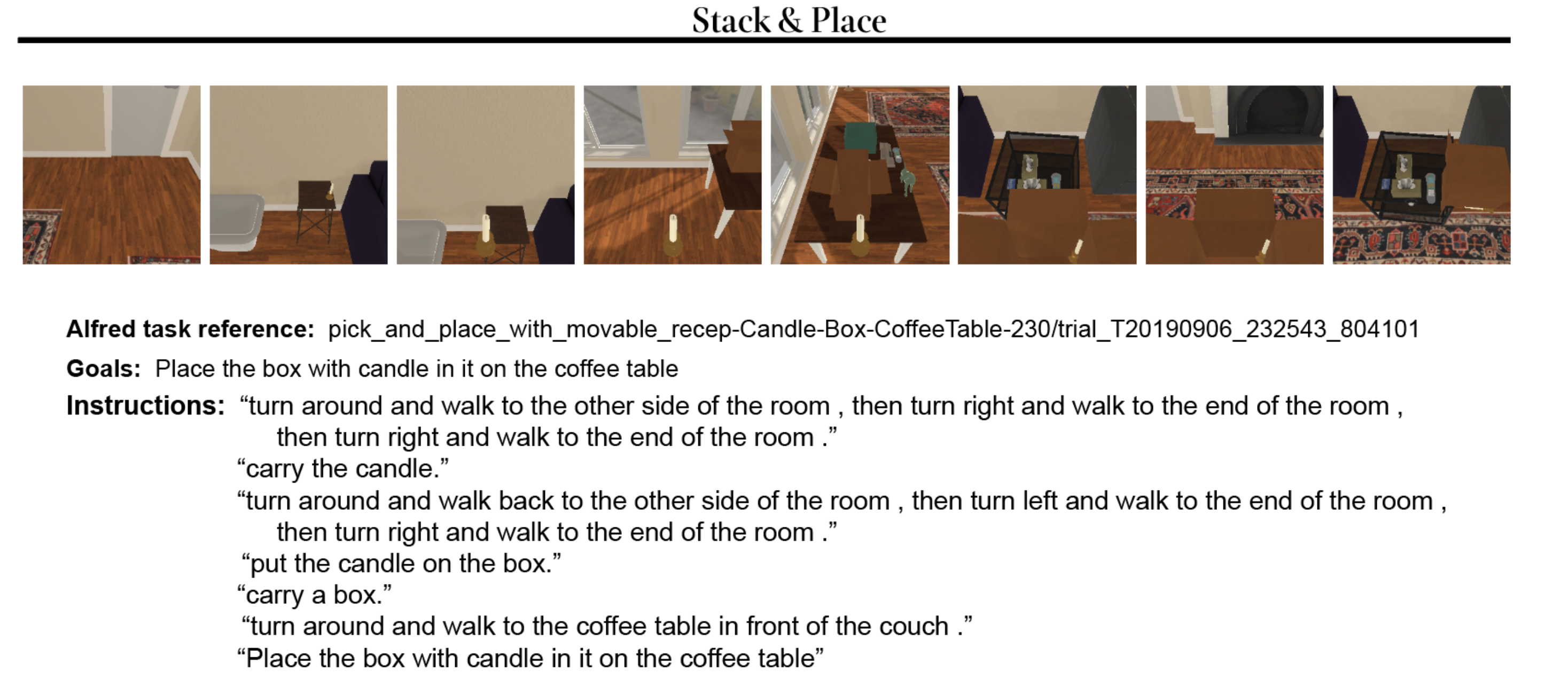}\\
    \includegraphics[width=0.95\textwidth]{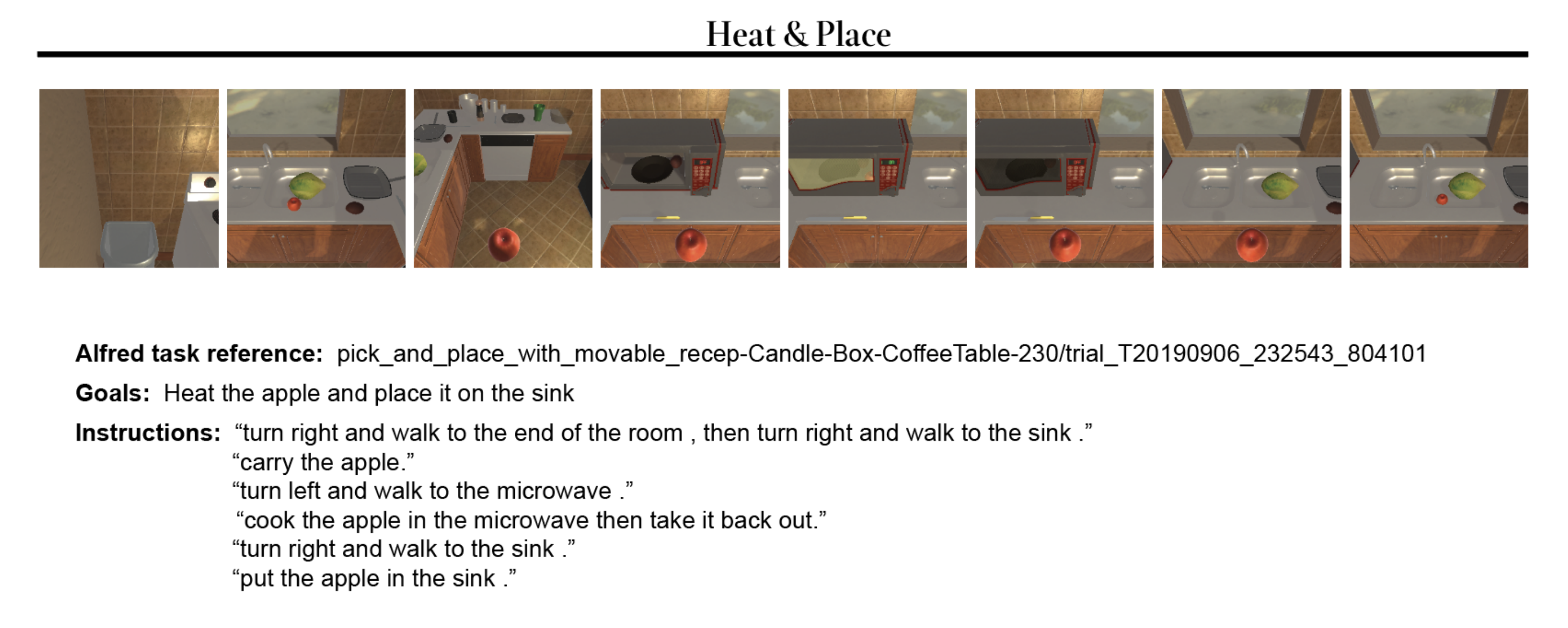}\\
    \includegraphics[width=0.95\textwidth]{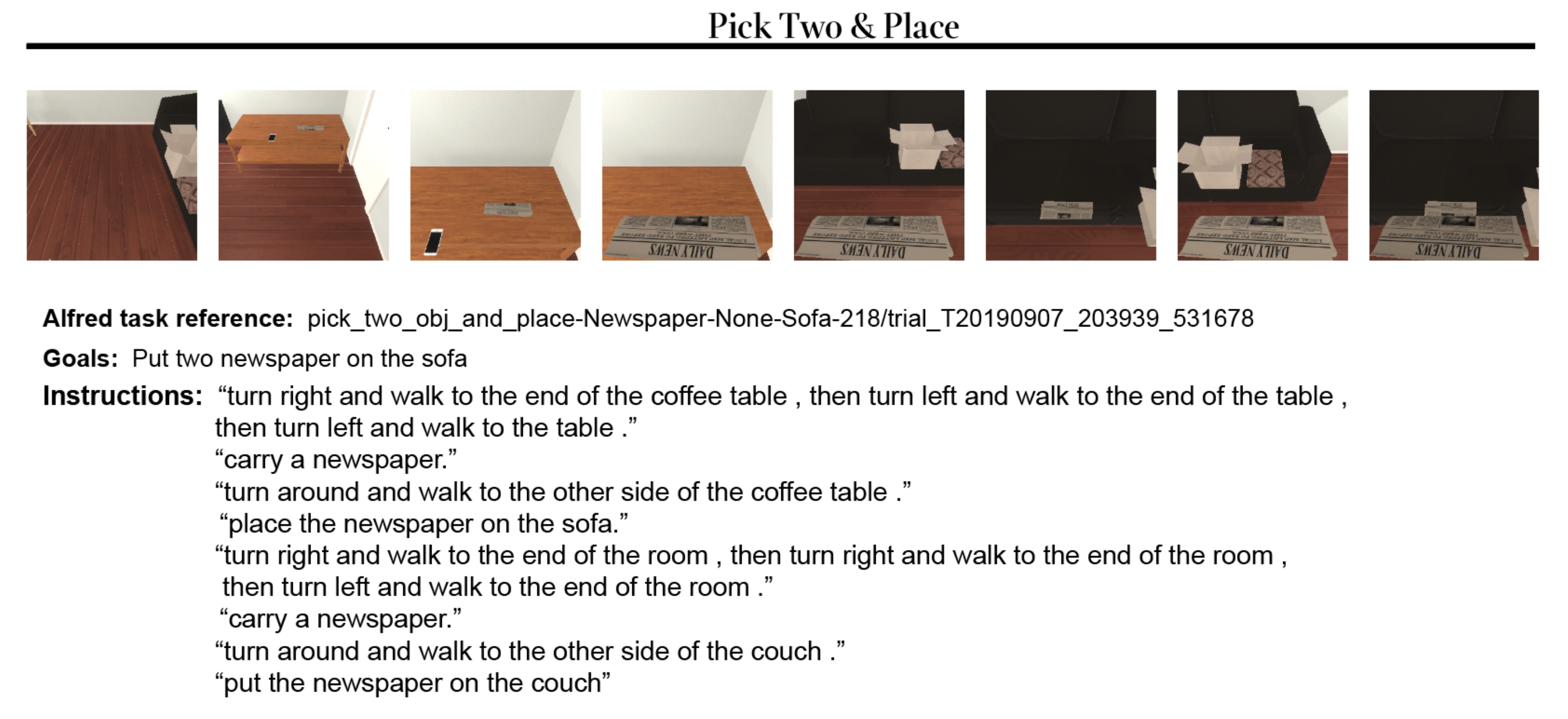}
\end{tabular}
    \caption{Dataset Examples. Automatically generated scenes, low-level actions, and language instructions.}
    \label{fig:my_label2}
\end{figure}

The dataset examples from \framework\ are shown in Figure~\ref{fig:my_label1} and Figure~\ref{fig:my_label2}.

\clearpage 


\vspace{-1em}
\begin{table}[h!]
\centering
\small
\begin{tabular}{c
>{\columncolor[HTML]{FFFFFF}}c 
>{\columncolor[HTML]{FFFFFF}}c 
>{\columncolor[HTML]{FFFFFF}}c cc}
\hline
\textbf{Simulator} & {\color[HTML]{222222} \textbf{\begin{tabular}[c]{@{}c@{}}Layout \\ randomization\end{tabular}}} & {\color[HTML]{222222} \textbf{\begin{tabular}[c]{@{}c@{}}Small Object \\ randomization\end{tabular}}} & {\color[HTML]{222222} \textbf{\begin{tabular}[c]{@{}c@{}}Object material \\ randomization\end{tabular}}} & \textbf{\begin{tabular}[c]{@{}c@{}}Number of \\ scenes/rooms\end{tabular}} & \textbf{\begin{tabular}[c]{@{}c@{}}Number of \\ objects\end{tabular}} \\ \hline
Habitat \\ (2020)     & {\color[HTML]{222222} \textbf{N/A}}                                                             & {\color[HTML]{222222} \textbf{N/A}}                                                                   & {\color[HTML]{222222} \textbf{N/A}}                                                                      & 120                                                                        & N/A                                                                   \\
Virtualhome \\ (2019) & {\color[HTML]{222222} \textbf{N/A}}                                                             & {\color[HTML]{222222} \textbf{N/A}}                                                                   & {\color[HTML]{222222} \textbf{\checkmark}}                                                & 7(house)                                                                   & 357                                                                   \\
threeDworld \\ (2021) & {\color[HTML]{222222} \textbf{N/A}}                                                             & {\color[HTML]{222222} \textbf{\checkmark}}                                             & {\color[HTML]{222222} \textbf{\checkmark}}                                                & 100+                                                                       & 1000+                                                                 \\
iGibson \\ (2021)     & {\color[HTML]{222222} \textbf{N/A}}                                                             & {\color[HTML]{222222} \textbf{\checkmark}}                                             & {\color[HTML]{222222} \textbf{N/A}}                                                                      & 106(house)                                                                 & 1984                                                                  \\
AI2Thor \\ (2021)     & {\color[HTML]{222222} \textbf{N/A}}                                                             & {\color[HTML]{222222} \textbf{\checkmark}}                                             & {\color[HTML]{222222} \textbf{\checkmark}}                                                & 227                                                                        & 2000+                                                                 \\ \hline
\textbf{Luminous}  & {\color[HTML]{222222} \textbf{\checkmark}}                                       & {\color[HTML]{222222} \textbf{\checkmark}}                                             & {\color[HTML]{222222} \textbf{\checkmark}}                                                & {\color[HTML]{222222} \textbf{$\infty$}}                                   & \textbf{2000+}                                                        \\ \hline
\end{tabular}
\newline
\caption{Comparison of \framework\ and existing embodied AI simulation platforms. Layout randomization specifies the simulator's ability to change the furniture layout; small object randomization refers to change the layout of items on an affordance such as table and countertop; object material randomization changes the texture and color of an object.}
\label{tab:simulations}
\end{table}

\end{document}